\documentclass[10pt,twocolumn,letterpaper]{article}

\usepackage[applications]{wacv}      

\definecolor{wacvblue}{rgb}{0.21,0.49,0.74}
\usepackage[pagebackref,breaklinks,colorlinks,allcolors=wacvblue]{hyperref}

\usepackage{censor} 
\usepackage{booktabs} 
\usepackage{tablefootnote} 
\usepackage{nicefrac} 
\usepackage{enumitem} 
 
\newcommand{\datasetname}{CSFD-1.6M{}}

\newif\ifdisablecdot
\disablecdottrue 

\let\origcdot\cdot

\ifdisablecdot
  \renewcommand{\cdot}{} 
  \newcommand{\ccdot}{\origcdot} 
\else
  \renewcommand{\cdot}{\origcdot} 
  \newcommand{\ccdot}{\origcdot}
\fi

\disablecdottrue

\def\wacvPaperID{64} 
\def\confName{WACV}
\def\confYear{2026}

\title{Photo Dating by Facial Age Aggregation}

\author{Jakub Paplh\'am\\
Czech Technical University in Prague\\
{\tt\small paplhjak@fel.cvut.cz}
\and
Vojtěch Franc\\
Czech Technical University in Prague\\
{\tt\small xfrancv@fel.cvut.cz}
}

\begin{document}
\maketitle
\begin{abstract}
We introduce a novel method for \texttt{Photo Dating} which estimates the year a photograph was taken by leveraging information from the faces of people present in the image. To facilitate this research, we publicly release \texttt{\datasetname}\footnote{Data and code will be published upon paper acceptance.}, a new dataset containing over 1.6 million annotated faces, primarily from movie stills, with identity and birth year annotations. Uniquely, our dataset provides annotations for multiple individuals within a single image, enabling the study of multi-face information aggregation. We propose a probabilistic framework that formally combines visual evidence from modern face recognition and age estimation models, and career-based temporal priors to infer the photo capture year. Our experiments demonstrate that aggregating evidence from multiple faces consistently improves the performance and the approach significantly outperforms strong, scene-based baselines, particularly for images containing several identifiable individuals.
\end{abstract}
\section{Introduction}
Determining when a photograph was taken is a task of considerable practical interest in computer vision. It provides significant value for organizing vast, chronologically unsorted digital archives where embedded metadata is often missing or unreliable. In fields such as digital forensics and journalism, estimating the capture date of an image is a powerful tool for verifying authenticity, helping to construct accurate event timelines and to detect manipulated or synthetically generated media by exposing temporal inconsistencies. The emergence of commercial tools aimed at the general public, such as the MyHeritage PhotoDater\textsuperscript{TM} \citep{myheritage2023photodater}, highlights the broader industrial interest in this problem. Current automated methods for image dating analyze general visual content. Researchers have developed models that learn temporal signals from era-specific objects such as cars and clothing, trends in photographic color processing, or overall visual style \citep{10.1007/978-3-030-86331-9_20, stacchio2022searching, 10.1007/978-3-319-56608-5_57, ginosar2017yearbooks, salem2016face2year, barancova2023blind}. While effective in certain scenarios, relying on such cues has limitations and overlooks what is arguably the most informative signal in many photographs: the human faces. The face provides a widely available and uniquely strong temporal signal, especially in contexts like genealogy and photo archiving, where analog scans often have absent or misleading metadata.

In this work, we introduce \textit{facial-age-based image dating}, a novel approach to the task of estimating the capture year of a photograph. Our method leverages the rich temporal information contained within human faces, provided at least one person in the image can be identified and their birth year is known. The core mechanism is to infer the capture year by adding the visually estimated age of a person to their known birth year. We propose a formal probabilistic framework that aggregates this evidence from all identifiable individuals to produce a single, robust date estimate. The framework formally combines visual evidence from state-of-the-art face recognition and age estimation approaches with career-based temporal priors, if available. To evaluate our method's core strength;   aggregating evidence from multiple people; existing benchmarks are unsuitable. We therefore construct \datasetname{}, a new large-scale dataset designed specifically for multi-person image dating. Derived from movie and TV show stills, it contains over 1.6 million annotated faces, each linked to a known identity, birth year, and an approximate image capture year.

\paragraph{Contributions:}
\begin{itemize}
\item We formalize a new \textit{face-based} method for image dating.
\item We release a new large-scale dataset, and show its value for both our method, and general age estimation.
\item We demonstrate that age-based evidence with career priors outperforms strong baselines, and that aggregating evidence from multiple faces improves accuracy.
\end{itemize}
\vspace{0.15cm}
The remainder of the paper is organized as follows. In \Cref{sec:related} we review the related work. \Cref{sec:model} presents our probabilistic model. \Cref{sec:dataset} describes the novel dataset constructed to evaluate the approach. \Cref{sec:experiments} reports experimental results, and \Cref{sec:conclusion} concludes with discussion.

\begin{figure*}
    \centering
    \begin{subfigure}[b]{0.47\linewidth}
        \centering
        \includegraphics[width=0.95\linewidth]{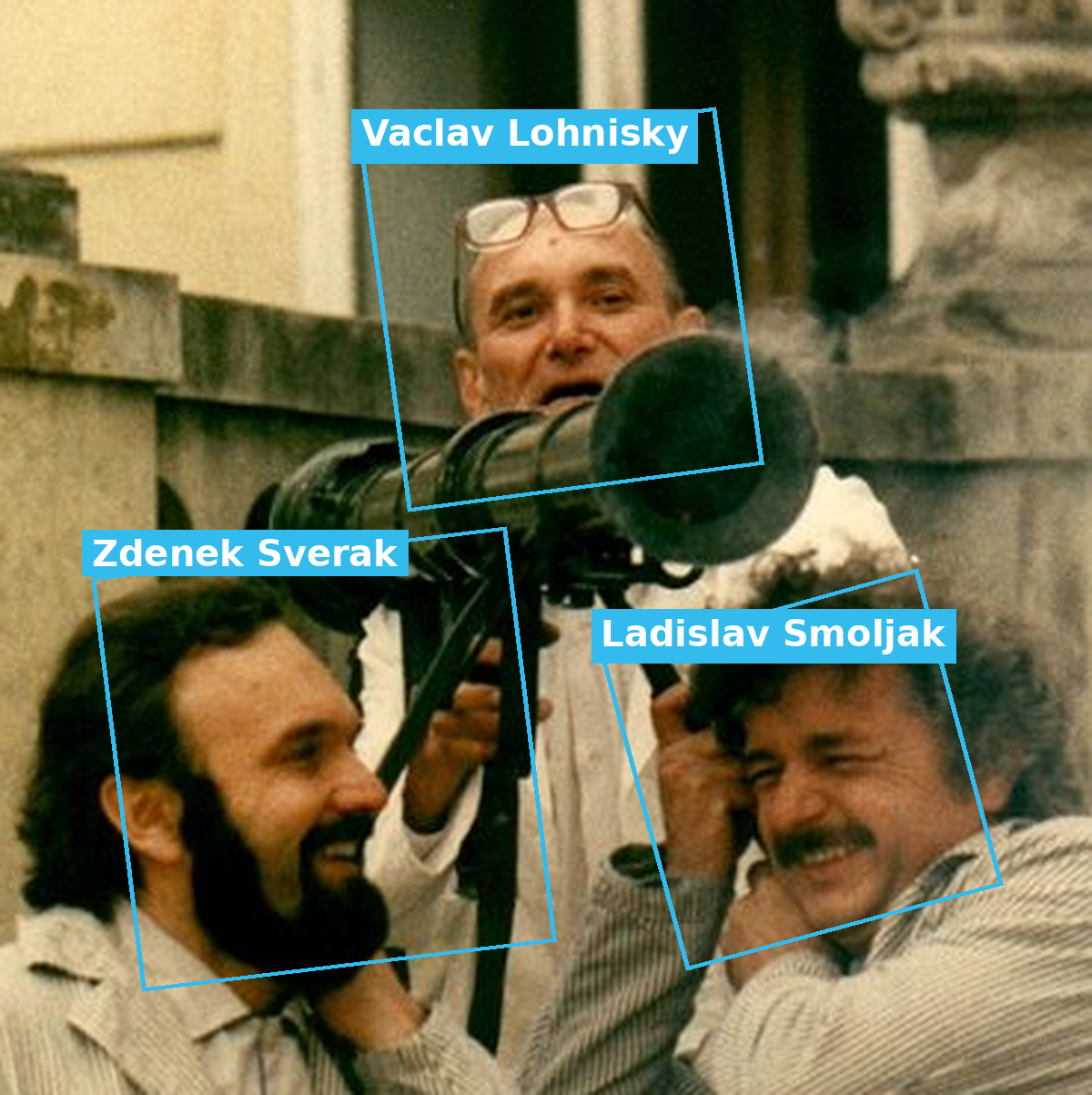}
        \caption{Image $x$ with detected faces $\mathbf{f} = (f_1, f_2, f_3)$ and the true identities. Still from \textit{Joachim, Put It in the Machine} (1974), \textcopyright~Barrandov Studio. Image included for non-commercial research purposes under fair use.}
        \label{fig:main_photo}
    \end{subfigure}%
    \hfill
    \begin{subfigure}[b]{0.43\linewidth}
        \centering
        \begin{subfigure}[t]{0.2\linewidth}
            \includegraphics[width=\linewidth]{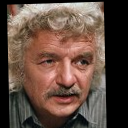}
            \vspace{0.4mm}
            
            \includegraphics[width=\linewidth]{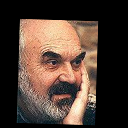}
            \vspace{0.4mm}
            
            \includegraphics[width=\linewidth]{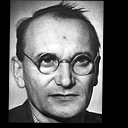}
            \caption*{Reference}
        \end{subfigure}
        \hfill
        \begin{subfigure}[t]{0.2\linewidth}
            \includegraphics[width=\linewidth]{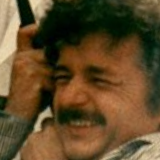}
            \vspace{0.4mm}
            
            \includegraphics[width=\linewidth]{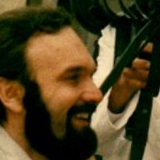}
            \vspace{0.4mm}
            
            \includegraphics[width=\linewidth]{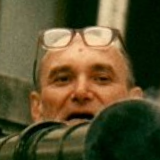}
            \caption*{Detected}
        \end{subfigure}
        \hfill
        \begin{subfigure}[t]{0.5\linewidth}
            \includegraphics[width=\linewidth]{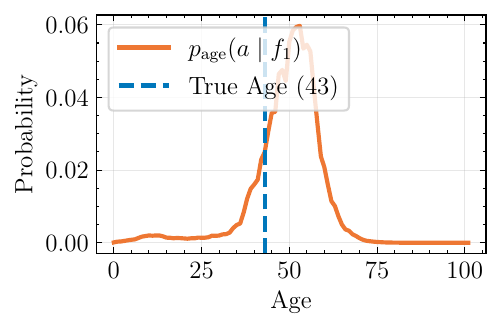}
            \includegraphics[width=\linewidth]{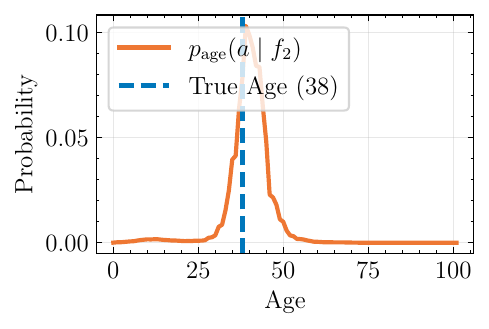}
            \includegraphics[width=\linewidth]{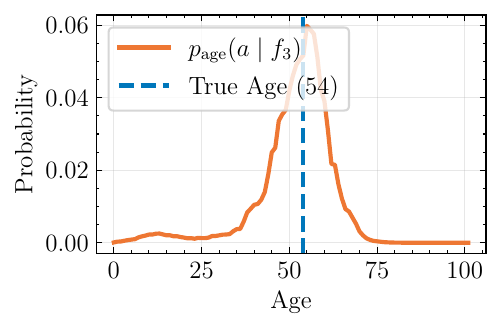}
        \end{subfigure}
        \caption{Recognized identities $\mathbf{i}=(i_1, i_2, i_3)$ in the detected faces $\mathbf{f}$. Knowing the birth year $b(i_j)$ for a recognized identity $i_j$ allows us to compute the year posterior $p(y \mid f_j, i_j) = p_{\text{age}}(a = y - b(i_j) \mid f_j)$ using the posteriors $p_{\text{age}}(a \mid f_j)$ from an age estimation model.}
        \label{fig:identity_components}
    \end{subfigure}

    \vspace{4mm} 

    \begin{subfigure}[b]{0.45\linewidth}
        \centering
        \includegraphics[width=0.9\linewidth]{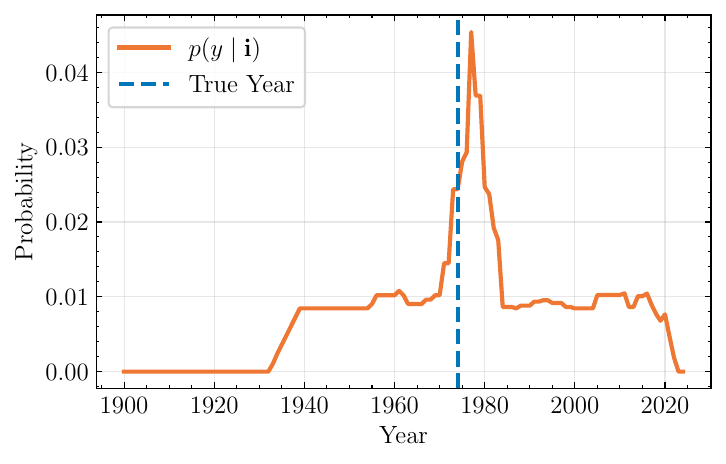}
        \caption{Joint career prior $p(y\mid \mathbf{i})$ for the recognized identities $\mathbf{i}$.}
        \label{fig:joint_prior}
    \end{subfigure}
    \hfill
    \begin{subfigure}[b]{0.45\linewidth}
        \centering
        \includegraphics[width=0.9\linewidth]{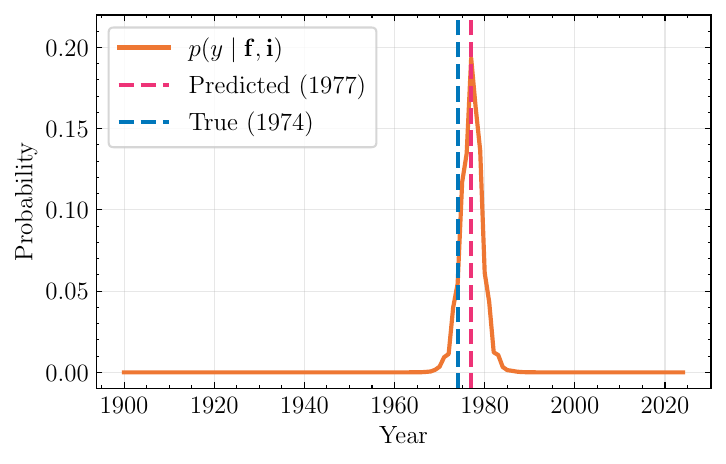}
        \caption{Final posterior distribution $p(y\mid \mathbf{f})$ of the image capture year.}
        \label{fig:final_posterior_dist}
    \end{subfigure}

    \caption{\textbf{Overview of the proposed dating framework.} Our method is demonstrated on an example image from \datasetname{}. 
    (a) The input image with detected faces and their ground-truth identities noted. 
    (b) For each detected face, we show the matched reference identity portrait and the corresponding year posterior, which is derived from the model's age estimate. 
    (c) The temporal information from the recognized identities is aggregated into a joint movie prior for the capture year. 
    (d) The final capture year posterior is obtained by combining the joint prior from (c) with the individual year posteriors from (b) using the model described in \Cref{eq:model_marginalization}.}
    \label{fig:final_layout}
\end{figure*}
\section{Related Work}
\label{sec:related}
Our research builds upon three distinct but related areas in computer vision: the general task of image capture date estimation, facial age estimation, and the use of contextual information in multi-person scenes. We review each in turn.
\vspace{-0.2cm}
\paragraph{Image Capture Date Estimation}
The task of automatically estimating a photograph capture year has been approached from several angles, primarily focusing on non-person-specific visual cues. One line of work learns temporal signals from the entire image, using general cues such as the physical properties of color film stocks \citep{10.1007/978-3-642-33783-3_36}. More recent methods \citep{10.1007/978-3-319-56608-5_57, 10.1007/978-3-030-86331-9_20, barancova2023blind} leverage large-scale datasets to train deep networks to find specific temporal cues in general visual content, such as vehicles or buildings. A second line of work uses human appearance as a temporal signal, but does so by modeling period-specific style. Salem \etal \citep{salem2016face2year} and Ginosar \etal \citep{ginosar2017yearbooks} use high school yearbooks to learn trends in clothing and hairstyles. In these approaches, the identity of any given person is irrelevant; their face and attire serve only as an anonymous sample of a collective, era-defining appearance. Our work fundamentally differs. We propose that the specific, unique identity of a person, when combined with their birth year, provides a much stronger and more direct temporal constraint than style alone.

\paragraph{Facial Age Estimation}
Estimating a person's age from a facial image is a core component of our model. This is a well-studied problem, with progress driven by large-scale datasets like IMDB-WIKI \citep{Rothe-IJCV-2018}. The dominant approach is to train deep neural network for age regression or classification \cite{Rothe-IJCV-2018, DLDL, DLDL-V2, MeanVariance, SoftLabels, OR-CNN, Paplham_2024_CVPR}. Despite significant progress, the recent NIST Face Analysis Technology Evaluation (FATE) reports that state-of-the-art models still exhibit considerable error, highlighting the difficulty of the task \citep{NIST_FRVT_AE}. Our work does not aim to advance age estimation itself; instead, we leverage a state-of-the-art age predictor as a probabilistic component, providing a distribution of ages to reason over.

\paragraph{Reasoning in Multi-Person Images}
While most face analyses treat faces in isolation, a few studies have explored using context from multiple people to improve performance on a given task. Li \etal \citep{7780514}, for instance, use the co-occurrence of other individuals in a photo to improve identity recognition. This is especially valuable when dealing with historical archives, where image quality poses a significant challenge \citep{mohanty2020photo}. These works use multi-person context to solve for identity. In contrast, our work leverages the identities and their apparent ages as joint evidence to infer a single, shared latent variable: the image capture year.
\section{Methodology}
\label{sec:model}
\subsection{Probabilistic Model}
\paragraph{Notation}
Let $x\in\mathcal{X}$ be an image taken in year $y\in\mathcal{Y}$ that contains a sequence $\mathbf{f} = (f_1, f_2, \ldots, f_n)$ of $n$ detected faces, where each $f_j\in\mathcal{F}$ is a crop extracted from $x$. Let $\mathcal{I} = \{\text{id}_1, \text{id}_2, \ldots, \text{id}_m\}$ be a finite set of $m\geq n$ identities. Let $\mathbf{i} = (i_1, i_2, \ldots, i_n)$ be an assignment of identities to the faces $\mathbf{f}$, where $i_j \in \mathcal{I}$ is the identity assigned to face $f_j$. As a person can appear at most once in an image, we enforce that all assigned identities are unique. Let $\mathcal{A}_n$ be the set of all valid identity assignments for the $n$ faces, $\mathcal{A}_n = \{ (i_1, \ldots, i_n) \in \mathcal{I}^n \mid i_j \neq i_k \text{ for } j \neq k \}$. Lastly, let $b:\mathcal{I}\rightarrow \mathbb{Z}$ be a mapping of each identity $\text{id} \in \mathcal{I}$ to their birth year.

\paragraph{Model}
Our goal is to derive the probability distribution $p(y\mid x)$ of the year an image $x$ was taken. We assume that the visual information relevant to the year $y$ is entirely encapsulated by the sequence of $n$ detected faces, $\mathbf{f}=(f_1, \ldots, f_n)$.
Therefore, we can write: $p(y\mid x) = p(y\mid \mathbf{f})$. The task then becomes to model and compute $p(y\mid \mathbf{f})$. We express the probability by marginalizing over all possible identity assignments $\mathbf{i} \in \mathcal{A}_n$ and using the chain rule, 
\begin{align}
\label{eq:model_marginalization}
p(y \mid  \mathbf{f}) = \sum_{\mathbf{i} \in \mathcal{A}_n} p(y, \mathbf{i} \mid  \mathbf{f}) = \sum_{\mathbf{i} \in \mathcal{A}_n} p(y \mid  \mathbf{f}, \mathbf{i}) \cdot p(\mathbf{i} \mid  \mathbf{f}).
\end{align}
The summation over the assignment set $\mathcal{A}_n$ can be computationally intractable if the candidate identity set $\mathcal{I}$ is large. We discuss an approach to managing this in \Cref{sec:experiments}.

\subsection{Simplifying Assumptions}
\paragraph{Modelling $p(\mathbf{i} \mid \mathbf{f})$}
For simplicity, we assume that given the identities $\mathbf{i} = (i_1, \ldots, i_n)$, the appearances of the face crops $\mathbf{f} = (f_1, \ldots, f_n)$ are conditionally independent, and the appearance of $f_j$ only depends on its identity $i_j$:
\begin{align} \label{eq:face_appearance_cond_indep}
p(\mathbf{f} \mid \mathbf{i}) = \prod_{j=1}^{n} p(f_j \mid i_j).
\end{align}
The assumption in \cref{eq:face_appearance_cond_indep} implies that factors such as shared lighting or camera parameters across faces in the same image are not explicitly modelled when the identities are known. This allows us to write $p(\mathbf{i} \mid \mathbf{f})$ for $\mathbf{i} \in \mathcal{A}_n$ as
\begin{align} 
\label{eq:posterior_assignment_orig}
p(\mathbf{i} \mid \mathbf{f}) &= \frac{p(\mathbf{f} \mid \mathbf{i}) \cdot p(\mathbf{i})}{p(\mathbf{f})} = \frac{\left(\prod_{j=1}^{n} p(f_j \mid i_j)\right) \cdot p(\mathbf{i})}{p(\mathbf{f})} \nonumber\\
&= \frac{\left(\prod_{j=1}^{n} p(f_j \mid i_j)\right) \cdot p(\mathbf{i})}{\sum_{\mathbf{i}' \in \mathcal{A}_n} \left(\prod_{k=1}^{n} p(f_k \mid i'_k)\right) \cdot p(\mathbf{i}')}.
\end{align}

\paragraph{Modelling $p(y\mid \mathbf{f}, \mathbf{i})$}

We assume conditional independence of face appearances given the year $y$ and identities $\mathbf{i}$, where each face $f_j$ depends only on $y$ and its identity $i_j$:
\begin{align} \label{eq:face_appearance_cond_indep_year}
p(\mathbf{f} \mid y, \mathbf{i}) = \prod_{j=1}^{n} p(f_j \mid y, i_j).
\end{align}
The term $p(y \mid \mathbf{f}, \mathbf{i})$ can then be expressed as:
\begin{align} 
\label{eq:bayes_y_f_i_initial}
p(y \mid \mathbf{f}, \mathbf{i}) &= \frac{p(\mathbf{f} \mid y, \mathbf{i}) \cdot p(y \mid \mathbf{i})}{p(\mathbf{f} \mid \mathbf{i})} \nonumber\\
&= \frac{\left(\prod_{j=1}^{n} p(f_j \mid y, i_j)\right) \cdot p(y \mid \mathbf{i})}{\sum_{y' \in \mathcal{Y}} \left(\prod_{k=1}^{n} p(f_k \mid y', i_k)\right) \cdot p(y' \mid \mathbf{i})}  \\
&= \frac{\left(\prod_{j=1}^{n} \frac{p(y \mid f_j, i_j) \cdot p(f_j \mid i_j)}{p(y \mid i_j)} \right) \cdot p(y \mid \mathbf{i})}{\sum_{y' \in \mathcal{Y}} \left(\prod_{k=1}^{n} \frac{p(y' \mid f_k, i_k)\cdot p(f_k \mid i_k)}{p(y' \mid i_k)} \right) \cdot p(y' \mid \mathbf{i})}. \nonumber
\end{align}

\paragraph{Note}
A point of consideration is the term $p(f_j \mid i_j)$, which appears in the formulation of both $p(\mathbf{i} \mid \mathbf{f})$ and $p(y \mid \mathbf{f}, \mathbf{i})$. In a strict generative model where year $y$ influences appearance per $p(f_j \mid y, i_j)$, the term $p(f_j \mid i_j)$ would ideally be derived as $\sum_{y' \in \mathcal{Y}} p(f_j \mid y', i_j)p(y' \mid i_j)$. Our current model treats $p(f_j \mid i_j)$ as a quantity that can be defined or estimated distinctly within each respective component. We acknowledge this pragmatic simplification.

\subsection{Implementation Choices}

The practical application of this model involves specific choices for its components, which we detail below.
\vspace{-0.2cm}
\paragraph{Year Posterior via Age Estimation}
The year estimation posterior $p(y \mid f_j, i_j)$ is derived from an age estimation model that outputs a distribution $p_{\text{age}}(a \mid f_j)$ over age $a$ given the face crop $f_j$. Knowing the birth year $b(i_j)$ for identity $i_j$, we set $p(y \mid f_j, i_j) = p_{\text{age}}(a = y - b(i_j) \mid f_j)$. To this end, we use a state-of-the-art age estimation model \texttt{cvut\_002}, which consists of a ViT-B/16 backbone trained with cross-entropy on a proprietary dataset, that ranks highly in the \href{https://pages.nist.gov/frvt/html/frvt_age_estimation.html}{NIST FATE} challenge \citep{NIST_FRVT_AE}. 

\paragraph{Face Likelihood via Recognition Embeddings}
The face likelihood $p(f_j \mid i_j)$ is modeled in the ArcFace \citep{deng2019arcface} embedding space. Let $\mathbf{e}(f_j) \in \mathbb{R}^{512}$ be the unit-norm embedding of face $f_j$. For an identity $\text{id}_k \in \mathcal{I}$, the likelihood is given by a von Mises-Fisher (vMF) distribution, $p(f_j \mid i_j=\text{id}_k)=C_D(\kappa) \exp(\kappa \boldsymbol{\mu}_k^T \mathbf{e}(f_j))$. The mean embedding $\boldsymbol{\mu}_k$ for each identity $\text{id}_k$ is the normalized average of its $N_k$ portrait embeddings, $\{\mathbf{e}_1, \dots, \mathbf{e}_{N_k}\}$, computed as $\boldsymbol{\mu}_k = \nicefrac{\sum_{i=1}^{N_k} \mathbf{e}_i}{\left\| \sum_{i=1}^{N_k} \mathbf{e}_i \right\|_2}$. The shared concentration parameter $\kappa$ is estimated from the same profile portraits. First, we generate a set of intra-identity cosine similarities using a leave-one-out scheme: for each identity with $N > 1$ portraits, we iteratively form a temporary prototype from $N-1$ embeddings and compute its similarity to the held-out embedding. The mean of all such similarities across all identities gives us the mean resultant length $\bar{R}$. We then estimate $\kappa$ using the approximation \citep{JMLR:v6:banerjee05a}, $\kappa = \nicefrac{\bar{R}(D - \bar{R}^2)}{1 - \bar{R}^2}$, where $D=512$ is the dimensionality of the embedding space.

\paragraph{Identity and Temporal Priors}
For the identity assignment prior $p(\mathbf{i})$, we use an uninformative uniform distribution over the set of all possible assignments $\mathcal{A}_n$. We model the joint temporal prior as the product of priors for individual identities, $p(y \mid \mathbf{i}) \propto \prod_{j=1}^{n} p(y \mid i_j)$. The strength and source of this prior knowledge can vary significantly in real-world scenarios. To evaluate the performance and the sensitivity to the prior choice of our framework, we define five distinct formulations for the individual priors $p(y\mid i_j)$:


\begin{enumerate}
\item \textbf{Uniform Prior $p_{\mathrm{U}}$}: A \textit{uniform} distribution over the 100 years after birth; the minimal-information scenario.
\item \textbf{Decade Prior $p_{\mathrm{D}}$}: Probability is uniform within a decade; probability mass per decade is proportional to image counts. A prior obtainable from external data.
\item \textbf{Movie Prior $p_{\mathrm{M}}$}: An empirical distribution of \textit{movies} per year in our dataset; a weak oracle prior for ablation.
\item \textbf{Image Prior $p_{\mathrm{I}}$}: An empirical distribution of \textit{images} per year in our dataset; a strong oracle prior for ablation.
\item \textbf{Combination Prior $p_{\mathrm{C}}$}: A convex \textit{combination} of the oracle and uniform priors, $p_{\mathrm{C}}= \lambda p_{\mathrm{I}}+ (1-\lambda)p_{\mathrm{U}}$, used to analyze the model sensitivity to prior strength.
\end{enumerate}

\subsection{Handling Unknown Identities (Open-Set Setup)}
In any real-world application, an image may contain unknown individuals or faces that cannot be confidently identified. Our framework can be extended to this open-set scenario by introducing a dedicated out-of-distribution (OOD) identity, $\text{id}_{\text{OOD}}$, into the identity set $\mathcal{I}$. This OOD identity is modeled with uninformative distributions: its face likelihood $p(f_j \mid i_j = \text{id}_{\text{OOD}})$ is uniform over the embedding space (a vMF distribution with $\kappa=0$), and the temporal prior $p(y \mid i_j = \text{id}_{\text{OOD}})$ and the temporal posterior $p(y \mid f_j, i_j = \text{id}_{\text{OOD}})$ are uniform over all years $\mathcal{Y}$. This allows the model to assign unknown faces to $\text{id}_{\text{OOD}}$, and prevents them from corrupting the final year estimate. It
is equivalent to ignoring low-confidence matches. The performance of this formulation is evaluated in \Cref{fig:open_set}. For other experiments, we operate in a closed-set setting.
\section{The \datasetname{} Dataset}
\label{sec:dataset}
To develop and evaluate our model, we collect a novel large-scale dataset for image dating, which we name \datasetname. The dataset's primary characteristic is its focus on images containing multiple individuals within a scene, each with an associated identity and temporal label. This is necessary for studying the contextual relationships central to our model, a task for which existing benchmarks, such as {IMDB-WIKI \citep{Rothe-IJCV-2018}}, are unsuitable as they are curated for single-face analysis, providing annotations for only one face per image.

\paragraph{Data Collection}
The dataset is constructed by scraping the Czecho-Slovak Movie Database (CSFD). We gathered high-resolution images, each linked to a specific production from which we extract a release year to serve as the temporal label. We acknowledge that the release is an approximation of the true capture date, a source of temporal uncertainty common to datasets in this domain. For each image, we also collect links to the unique profiles of all pictured public figures, providing their birth years and portraits.

\begin{figure*}
  \centering
    \begin{subfigure}{0.3\linewidth}
    \includegraphics[width=\linewidth]{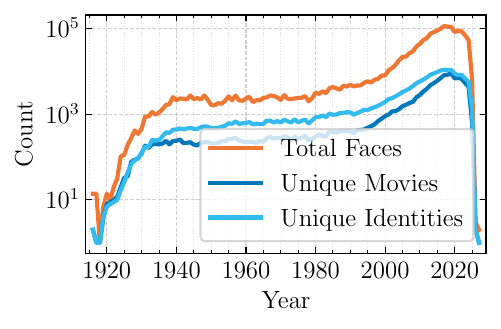}
    \caption{Temporal distribution of the data}
    \label{fig:short-c}
  \end{subfigure}
  \hfill
    \begin{subfigure}{0.3\linewidth}
    \includegraphics[width=\linewidth]{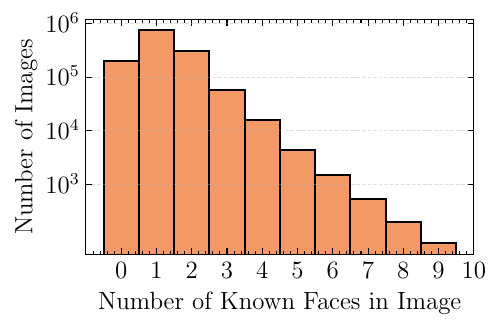}
    \caption{Distribution of known faces per image}
    \label{fig:short-b}
  \end{subfigure}
  \hfill
  \begin{subfigure}{0.3\linewidth}
    \includegraphics[width=\linewidth]{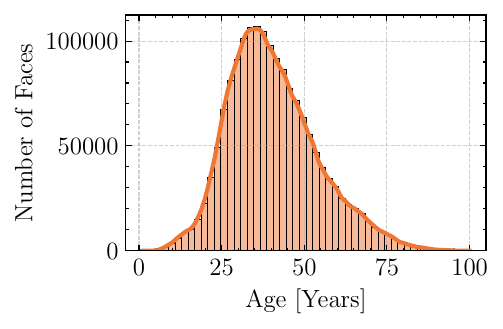}
    \caption{Distribution of face ages}
    \label{fig:short-a}
  \end{subfigure}
  \caption{\textbf{Overview of the \datasetname{} dataset statistics.}}
  \label{fig:dataset_statistics}
\end{figure*}

\paragraph{Identity Annotation Pipeline}
We use an automated pipeline to assign detected faces to the known identities linked to an image. For an image with $n$ faces detected by SCRFD-10GF \citep{guo2022sample} and $k$ linked identities:
\begin{enumerate}
    \item \textbf{Representation:} We compute ArcFace \citep{deng2019arcface} embeddings for all faces. A prototype embedding for each identity is created by averaging embeddings from their profile portraits, following the standard practice \citep{deng2019arcface}.
    \item \textbf{Constrained Assignment:} The assignment between the $n$ faces and $k$ identities is solved using the Hungarian algorithm on a cosine similarity matrix. We employ a standard formulation that pads the cost matrix to handle unbalanced cases $n\neq k$ and allows for partial matching\footnote{We confirmed the robustness of this pipeline by manually annotating a stratified sample of 1000 faces from images containing $1$ to $11$ faces. The algorithm correctly found $997$ matches and produced no incorrect matches.}. To ensure high-quality matches, costs for invalid pairings are set to infinity. A pair is invalid if its similarity is below $0.2$ or if the true age differs from the visually estimated age by more than $20 + 0.25 \cdot \text{age}_{\text{est}}$ years, allowing for greater error in age estimation for older subjects\footnote{Removing the age constraint yielded only $1\%$ additional matches on the full dataset, which we postulate are of lower quality.}.
\end{enumerate}
\vspace{0.1cm}
\noindent We do not filter faces by attributes such as size or resolution, leaving such preprocessing to downstream applications.
\vspace{-0.25cm}
\paragraph{Dataset Statistics}
The pipeline results in a dataset containing over 1.6 million faces, each with an assigned identity, birth year, and image year. Its key characteristic are multi-person scenes. Statistics of the dataset are detailed in \Cref{tab:dataset_stats}, \Cref{fig:dataset_statistics} and the supplementary material.

\begin{table}[]
\centering
\begin{tabular}{lr}
\textbf{Statistic} & \textbf{Value} \\
\midrule
Total Scraped Images & 1,493,357 \\
Total Detected Faces & 3,655,604 \\
Faces Matched to an Identity & 1,921,164 \\
\textbf{Final Faces (Matched \& Dated)} & 
\textbf{1,653,871} \\
\textbf{Final Images} & \textbf{1,143,213} \\
\midrule
Total Annotatable Identities & 75,952 \\
Unique Identities in Final Set\tablefootnote{Many (29.429) of the 75,952 valid profiles have no gallery images.} & 46,523 \\
\bottomrule
\end{tabular}
\caption{\textbf{Statistics of the \datasetname{} creation pipeline.}}
\label{tab:dataset_stats}
\end{table}
\paragraph{Comparison to Existing Datasets}
\Cref{tab:dataset_comparison} contrasts \datasetname{} with IMDB-WIKI, its closest large-scale analogue. Our dataset is not only substantially larger but, critically, consists of multi-person scenes required by our model. While no web-scraped dataset is free of label noise, our annotation pipeline leverages the explicit links between images and identity profiles provided by the database, aiming for higher-quality identity assignments. Other datasets designed for age estimation are discussed in \Cref{subsec:age_estimation}.
\begin{table}[]
\centering
\begin{tabular}{l c c}
\textbf{Attribute} & \textbf{IMDB-WIKI} & \textbf{\datasetname} \\
\midrule
Size (Faces) & $\sim$523k & $\sim$1.6M \\
Identities & $\sim$20k & $\sim$46k \\
Multi-face & Single-face annotations & Core feature \\
\bottomrule
\end{tabular}
\caption{\textbf{Comparison of \datasetname{} with IMDB-WIKI \citep{Rothe-IJCV-2018}.}\vspace{-0.3cm}}
\label{tab:dataset_comparison}
\end{table}
\paragraph{Public Release}
We release the \datasetname{} annotations to support further research in this area. The release includes all annotations: image URLs, face bounding boxes, identity assignments, and birth years. We also provide the pre-computed ArcFace embeddings and per-face age posterior distributions. This allows the research community to focus directly on modeling the contextual information shared between faces, as proposed here, without the significant overhead of data collection and feature extraction.
\section{Experiments and Results}
\label{sec:experiments}

In this section we present the experimental evaluation of our proposed model. We first describe the baseline and oracle models used for comparison, then detail the evaluation protocol, and finally present and discuss the results.

\subsection{Compared Models}
We evaluate four distinct models designed to ablate different components of our probabilistic framework. This allows us to quantify the impact of face identification errors, the assignment strategy, and the inclusion of temporal priors.

\paragraph{Oracle Model}
This model serves as an \textit{oracle} for our task. It assumes perfect knowledge of the identities in the image. For a given image with faces $\mathbf{f}$, we use the single ground-truth identity assignment $\mathbf{i}_{\text{GT}}$. This model isolates the performance of the age estimation and temporal priors, completely removing any error from the face identification.

\paragraph{Full Probabilistic Model (Full)}
This model is a practical implementation of the framework proposed in \Cref{sec:model}. It approximates the posterior $p(y \mid \mathbf{f})$ by marginalizing over a tractable subset of all identity assignments. This subset is constructed from candidate pools that cover $99\%$ of the probability mass for each face, capturing the most plausible joint assignments yet remaining computationally feasible.

\paragraph{Top-1 Assignment Model (Top-1)}
This model is a computationally efficient approximation of the full model. Instead of marginalizing, it determines the single \textit{joint} identity assignment $\mathbf{i}_{\text{Top-1}}$ that has the highest posterior probability, $\mathbf{i}_{\text{Top-1}} = \arg\max_{\mathbf{i} \in \mathcal{A}_n} p(\mathbf{i} \mid \mathbf{f})$. The final year distribution is then calculated conditioned entirely on this single assignment. This differs from a simple greedy approach as it finds the most likely valid assignment for the group as a whole.

\paragraph{Naive Baseline (Naive)}
This model serves as a simple \textit{baseline} to demonstrate the value of the probabilistic framework. It performs a greedy, \textit{independent} identification for each face, assigning it to the single best-matching identity regardless of the other face assignments. It then calculates the final year distribution by taking the product of the individual year posteriors from the age estimator, $p(y \mid \mathbf{f}) \propto \prod_{j=1}^{n} p(y \mid f_j, i_j)$, notably without the use of any temporal prior $p(y \mid \mathbf{i})$. This model reflects a straightforward but theoretically less grounded approach to the problem.


\paragraph{Scene-Based Baseline (Scene)}
To benchmark the face-based approach against the established state-of-the-art, we implement the direct prediction paradigm \citep{10.1007/978-3-319-56608-5_57, 10.1007/978-3-030-86331-9_20, barancova2023blind}, modeling $p(y\mid x)$ from the full scene context. 
This paradigm has been shown to be more effective than alternative approaches like image retrieval \citep{10.1007/978-3-030-86331-9_20}. 
Our Scene baseline consists of a ViT-B/16\footnote{Identical architecture to the age estimation model.} backbone that takes an entire image as input and outputs a probability distribution over the years.\footnote{We did not benchmark the older method \citep{10.1007/978-3-642-33783-3_36} as it is not competitive.} Unlike the face-based methods, which leverage pre-trained models, this baseline is trained directly on \datasetname{}, as detailed in the protocol below. This represents a best-case scenario for the \textit{Scene} model and a strong baseline.

\subsection{Evaluation Protocol}
We evaluate the models using the \textit{Mean Absolute Error} (MAE) between the predicted year and the ground-truth year. For all evaluated models the final year prediction is derived as the median of the posterior distribution $p(y \mid \ccdot )$.

\paragraph{Dataset Splits}
For the \textit{Scene} baseline, we perform 5-fold cross-validation. For each fold, we use a 60/20/20 train/validation/test split. The splits are created to be \textit{movie-disjoint}\footnote{An identity-disjoint split was not feasible, as the high interconnectedness of actors clustered the vast majority of the dataset into a single split.}, ensuring that all images from the same movie belong to only one split. We report the mean MAE across the 5 folds for the \textit{Scene} baseline. In contrast, the face-based models are trained on external datasets. Therefore, they are evaluated on the entirety of the \datasetname{} dataset.

\paragraph{Scene Baseline Training}
The \textit{Scene} model is trained for 200 epochs using the Adam optimizer ($\text{lr}=10^{-4}$), employing early stopping based on the validation set MAE with a patience of 10 epochs. The input images are resized so their shorter side is 384 pixels, from which a 384x384 crop is taken. For training, we apply standard data augmentations.

\subsection{Photo Dating Results}
This section presents the quantitative results of our experiments. Our central finding is that the face-based models consistently outperform the strong, in-domain \textit{Scene} baseline for images containing more than one person, validating the hypothesis that aggregating facial information (when available) is a superior strategy to general scene analysis.

\begin{figure}
  \centering
    \includegraphics[width=0.89\linewidth]{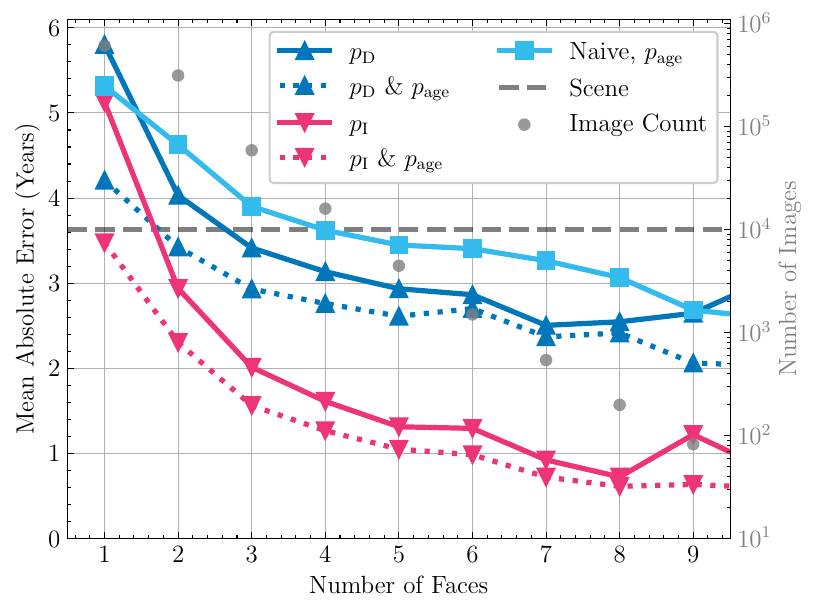}
    \caption{\textbf{Ablation of information sources.} MAE of predictions made using only a temporal prior ($p_{\text{D}}$ or $p_{\text{I}}$), only the age posterior ($p_{\text{age}}$, equivalent to the \textit{Naive} model), and the full model combining both ($p_{\text{D}}\,\&\, p_{\text{age}}$ or $p_{\text{I}}\,\&\, p_{\text{age}}$). The results demonstrate that both the temporal prior and the age posterior are informative signals. Combining them using the proposed aggregation consistently yields lower error than using either source in isolation.\vspace{-0.3cm}}
    \label{fig:prior_and_posterior_components}
\end{figure}

\paragraph{Ablation of Information Sources}
To understand the relative importance of the two primary signals in our model; the age-based posteriors $p(y\mid i_j, f_j)$ and the temporal prior $p(y\mid \mathbf{i})$; we conduct an ablation study shown in \Cref{fig:prior_and_posterior_components}. We evaluate the performance using only the temporal prior and only the aggregated age posteriors (equivalent to our \textit{Naive} baseline). The results show that both components are individually informative. Crucially, the full model, which combines both signals, consistently outperforms using either source in isolation. For real-world applications where a specific temporal prior may be unknown or difficult to obtain, our results validate that the age-based posterior alone still serves as a strong, standalone temporal signal. The full framework provides a principled way to further improve this baseline estimate whenever a prior becomes available.

\begin{figure}
  \centering
    \includegraphics[width=0.92\linewidth]{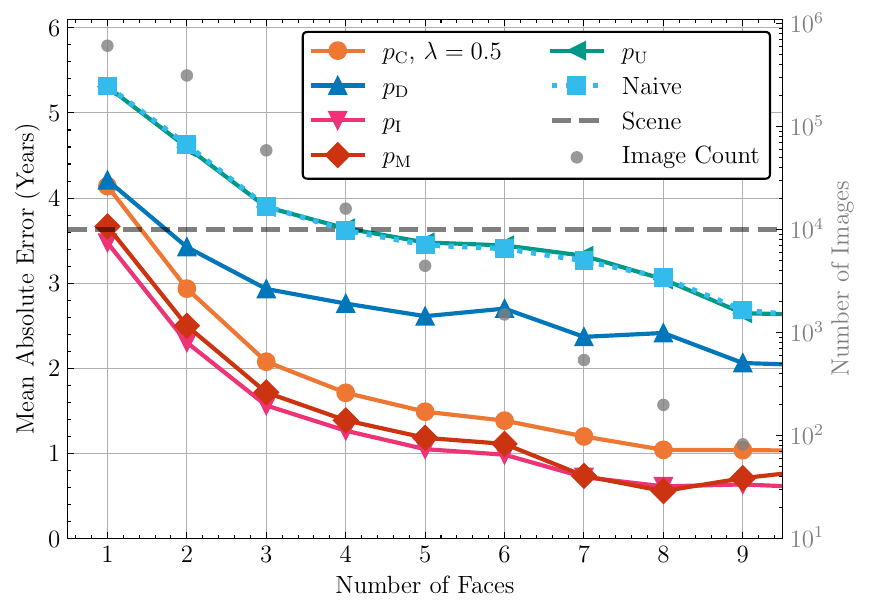}
    \caption{\textbf{Performance comparison of temporal priors.} MAE for the \textit{Full} model using different priors, and MAE of the \textit{Naive} and \textit{Scene} baselines. The results highlight the benefit of informative priors. The realistic \textit{Decade Prior} ($p_{\mathrm{D}}$) consistently outperforms the \textit{Naive} baseline and the uninformative \textit{Uniform Prior} ($p_{\mathrm{U}}$). The oracle priors ($p_{\mathrm{I}}$, $p_{\mathrm{M}}$, $p_{\mathrm{C}}$), which use statistics from the test set, are included to illustrate an upper bound on performance.\vspace{-0.15cm}}
    \label{fig:prior_ablation}
\end{figure}

\paragraph{Impact of Temporal Priors}
\Cref{fig:prior_ablation} details the effect of different temporal priors on the \textit{Full} model. The results demonstrate the value of using an informative prior. The realistic \textit{Decade Prior} ($p_{\mathrm{D}}$) provides a significant improvement over both the \textit{Naive} model (which uses no temporal prior) and the uninformative \textit{Uniform Prior} ($p_{\mathrm{U}}$). Notably, while the \textit{Scene} baseline is superior for single-face images, the \textit{facial-age-based} approach surpasses it for all images containing two or more faces. This occurs despite the \textit{Scene} model being trained directly on this dataset.

\begin{figure}
  \centering
    \includegraphics[width=0.9\linewidth]{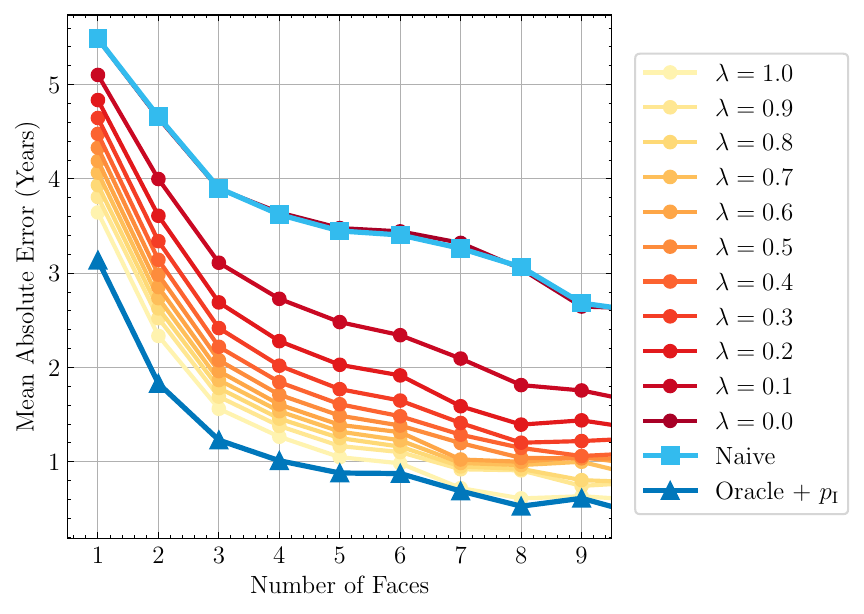}
    \caption{\textbf{Impact of prior strength ($\lambda$) on model performance.} MAE of the \textit{Full} model using the combination prior $p_{\mathrm{C}}$, as $\lambda$ interpolates between a purely uniform prior ($\lambda=0$) and the full oracle image prior ($\lambda=1$). The results show that even weak prior information yields substantial performance gain over the baseline.\vspace{-0.15cm}}
    \label{fig:lambda_ablation}
\end{figure}

\begin{figure}
  \centering
    \includegraphics[width=0.9\linewidth]{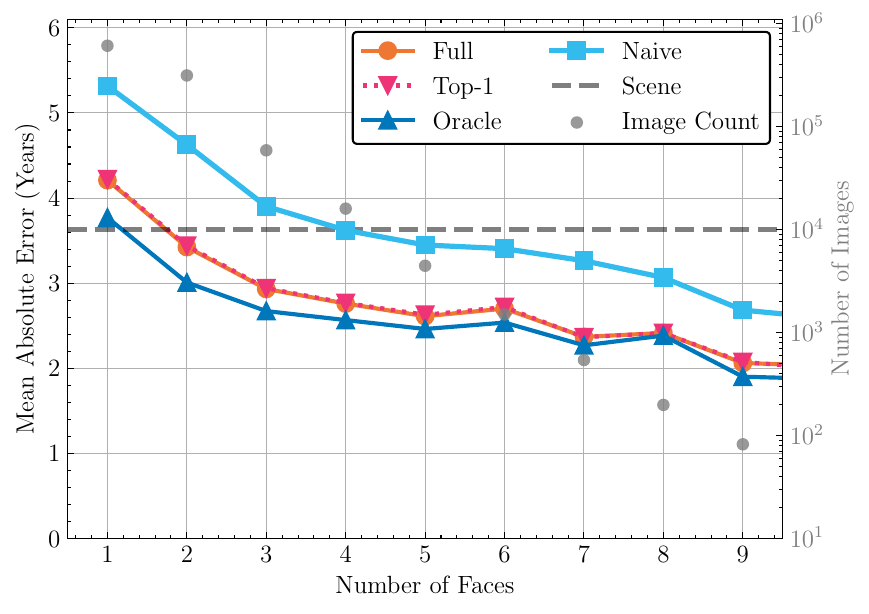}
    \caption{\textbf{Ablation of model components using a realistic prior.} MAE of the \textit{Oracle}, \textit{Full}, and \textit{Top-1} models with a decade prior ($p_{\mathrm{D}}$), compared against the \textit{Naive} and \textit{Scene} baselines. The \textit{Full} and \textit{Top-1} models outperform the strong \textit{Scene} baseline, demonstrating the power of aggregating evidence from multiple faces even when restricted to the weak realistically obtainable prior $p_{\text{D}}$.\vspace{-0.25cm}}
    \label{fig:model_ablation}
\end{figure}

\begin{figure}
  \centering
    \includegraphics[width=0.8\linewidth]{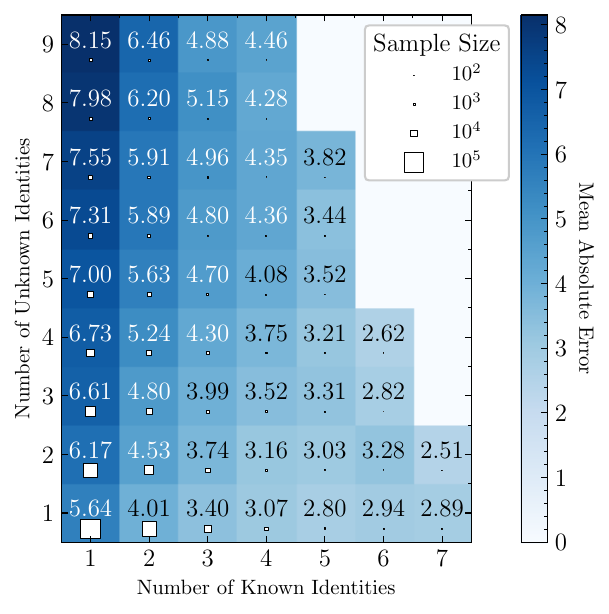}
    \caption{\textbf{Open-set evaluation.} MAE of the \textit{Full} model using prior $p_{\text{D}}$ stratified on the number of \textit{known} and \textit{unknown} identities.\vspace{-0.2cm}}
    \label{fig:open_set}
\end{figure}

\begin{figure}
  \centering
    \includegraphics[width=0.9\linewidth]{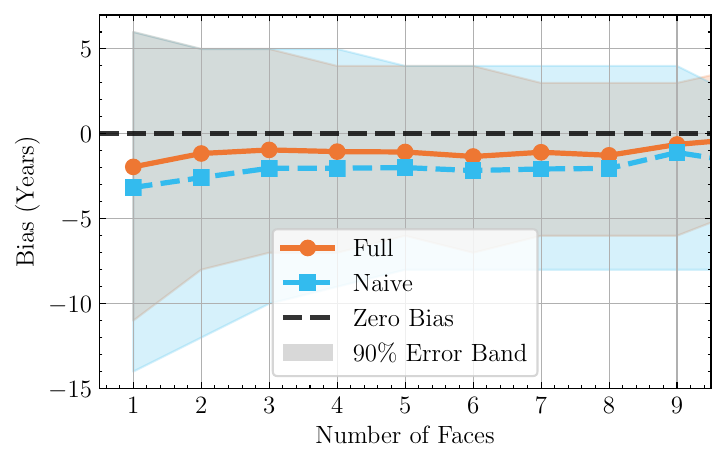}
    \caption{\textbf{Analysis of prediction bias.} Bias (Predicted - True) for the \textit{Full} model using the $p_{\mathrm{D}}$ prior, and the \textit{Naive} model. For both models we show a shaded band representing the central 90\% of prediction errors. Both models exhibit a consistent negative bias.\vspace{-0.3cm}}
    \label{fig:bias_analysis}
\end{figure}

\paragraph{Sensitivity to Prior Strength}
We analyze the model sensitivity to prior information in \Cref{fig:lambda_ablation}. The plot shows the MAE for the \textit{Full} model using the combination prior $p_{\mathrm{C}}$, as $\lambda$ interpolates from a purely uniform prior ($\lambda=0$) to the oracle image prior ($\lambda=1$). At $\lambda=0$, performance matches the \textit{Naive} baseline. As $\lambda$ increases, performance steadily improves, demonstrating that the framework effectively leverages prior information. Even a weak prior signal (e.g., $\lambda=0.1$) provides a substantial performance gain.
\vspace{-0.14cm}
\paragraph{Ablation of Marginalization}
To understand the contribution of each part of our framework we ablate the components in \Cref{fig:model_ablation} using the realistic decade prior ($p_{\mathrm{D}}$) for all face-based models. The \textit{Oracle} model establishes a clear performance bound. The key finding is that our \textit{Full} model, which marginalizes over possible identity assignments, offers only a marginal improvement over the much simpler \textit{Top-1} approximation. This suggests that the assignment posterior is sharply peaked, likely due to the high quality of the ArcFace embeddings, making the \textit{Top-1} approach a highly effective and computationally tractable alternative.
\vspace{-0.14cm}
\paragraph{Analysis of Prediction Bias}
We analyze the systematic bias of our predictions in \Cref{fig:bias_analysis}. The plot shows the distribution of the prediction error (Predicted Year - True Year). The models exhibit a consistent negative bias, predicting capture dates that are on average a few years earlier than the release year. This offset is an expected artifact of our data source and reflects the inherent lag between the film production (photos are taken) and its eventual release.
\vspace{-0.14cm}
\paragraph{Robustness to Unknown Identities}
Real-world images often contain a mix of known and unknown individuals. We evaluate this open-set setup in \Cref{fig:open_set}, which shows how the MAE is affected by the number of \textit{known} and \textit{unknown} faces. 
The error consistently decreases as the number of known identities increases. Conversely, while the error increases with more unknown identities, this degradation is gradual. This confirms that the framework is effective at preventing unknown faces from corrupting the estimate.

\begin{table}[t]
\centering
\small
\begin{tabular}{@{}l ccc@{}}
& \multicolumn{3}{c}{\textbf{Pre-training Dataset}} \\
\cmidrule(l){2-4} 
\textbf{Benchmark} & \textbf{ImageNet} & \textbf{IMDB-WIKI} & \textbf{\datasetname{}} \\
\midrule
AgeDB      & 7.05 $\pm$ 0.29 & 6.34 $\pm$ 0.25 & \textbf{5.25 $\pm$ 0.21} \\
AFAD       & 3.19 $\pm$ 0.04 & 3.10 $\pm$ 0.03 & \textbf{3.04 $\pm$ 0.03} \\
MORPH      & 2.98 $\pm$ 0.05 & 2.88 $\pm$ 0.07 & \textbf{2.76 $\pm$ 0.05} \\
UTKFace    & 4.84 $\pm$ 0.08 & 4.64 $\pm$ 0.06 & \textbf{4.08 $\pm$ 0.03} \\
CLAP2016   & 5.87 & 4.89 & \textbf{3.52} \\
\bottomrule
\end{tabular}
\caption{\textbf{Age estimation results.} MAE $\downarrow$ ($\pm$ std) on five benchmarks after pre-training a ResNet-101 model on different datasets.\vspace{-0.25cm}}
\label{tab:age_estimation_results}
\end{table}

\subsection{Age Estimation Performance}
\label{subsec:age_estimation}
To validate \datasetname{} as a high-quality resource, we evaluate its effectiveness as a pre-training corpus for age estimation. Following the protocol from \citep{Paplham_2024_CVPR}, we compare a ResNet-101 model initialized with ImageNet weights \citep{ImageNet} to models further pre-trained on either IMDB-WIKI \citep{Rothe-IJCV-2018} or \datasetname{}. Each model is then fine-tuned and evaluated on five datasets: AgeDB \citep{AgeDB}, AFAD \citep{AFAD}, MORPH \citep{MORPH}, UTKFace \citep{UTKFace}, and CLAP2016 \citep{CLAP}. The resulting MAE is reported in \Cref{tab:age_estimation_results}. The results demonstrate that pre-training on \datasetname{} consistently and substantially outperforms pre-training on ImageNet and IMDB-WIKI across all five evaluation benchmarks. This validates \datasetname{} not only for the proposed dating task but also as a valuable resource for the broader task of facial age estimation.

\section{Conclusion}
\label{sec:conclusion}
In this work, we introduced a novel method of \textit{Photo Dating by Facial Age Aggregation}, demonstrating that the faces of individuals within an image are a powerful source of temporal information. We presented a probabilistic model to formally combine visual evidence from the apparent age and career-based temporal priors. A key contribution of our research is the creation and public release of \datasetname{}, a large-scale dataset of over 1.6 million annotated faces, specifically designed to facilitate the study of multi-face reasoning. Our experiments validate our approach. We have shown that aggregating evidence from multiple faces consistently improves dating accuracy, outperforming a strong model based on current state-of-the-art \citep{10.1007/978-3-319-56608-5_57, 10.1007/978-3-030-86331-9_20, barancova2023blind}. Our work provides a foundation for future research into face-based image dating and temporal analysis of media.

\paragraph{Limitations \& Ethical Considerations} The probabilistic model, detailed in \Cref{sec:model}, relies on simplifying assumptions for tractability. The dataset is limited by the inherent temporal noise from using movie release years as capture dates. Some temporal priors are derived from the dataset itself, although our method also functions with weak and uninformative priors.
We release our contributions to foster research in temporal media analysis and encourage their responsible application, however, we recognize the associated ethical duties. The use of public data scraped without explicit consent, the demographic bias and the potential for dual-use are significant concerns. To mitigate these while supporting reproducibility, we don't distribute the images, but only the annotations, pre-computed features and URLs.

{
    \small
    \bibliographystyle{ieeenat_fullname}
    \bibliography{main}
}

\clearpage
\section*{Supplementary Material}
\section*{Stratified Performance Analysis}
\begin{figure*}
  \centering
  \includegraphics[width=0.9\linewidth]{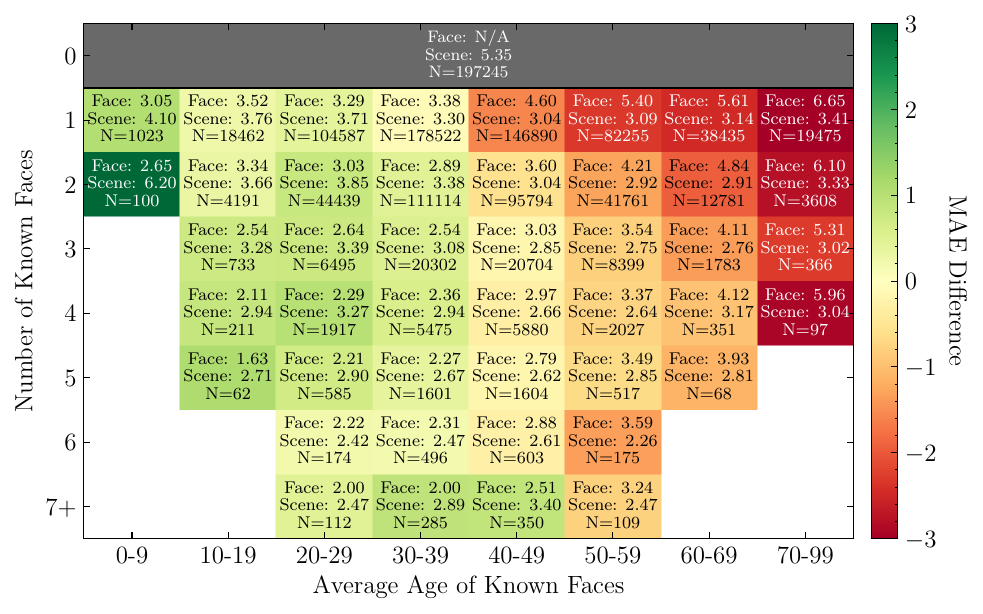}
    \caption{\textbf{Stratified performance.} MAE $\downarrow$ on CSFD-1.6M of the \textit{Face} ($p_{\text{D}}$) and \textit{Scene} methods, where $N$ denotes the sample-size. The results reflect that age-estimation is more precise for younger subjects.}
    \label{fig:error_by_age_big}
\end{figure*}

\Cref{fig:error_by_age_big} presents a detailed performance analysis of our face-based model against the \textit{Scene} baseline, stratified by both the number of faces and the average age of subjects. The analysis reveals that while the performance of the \textit{Scene} baseline is higher for older subjects, our face-based method is significantly more effective for images of younger individuals. This performance advantage for our method further increases with the number of identifiable faces in an image.

\begin{figure*}
    \centering
    \includegraphics[height=3.2cm]{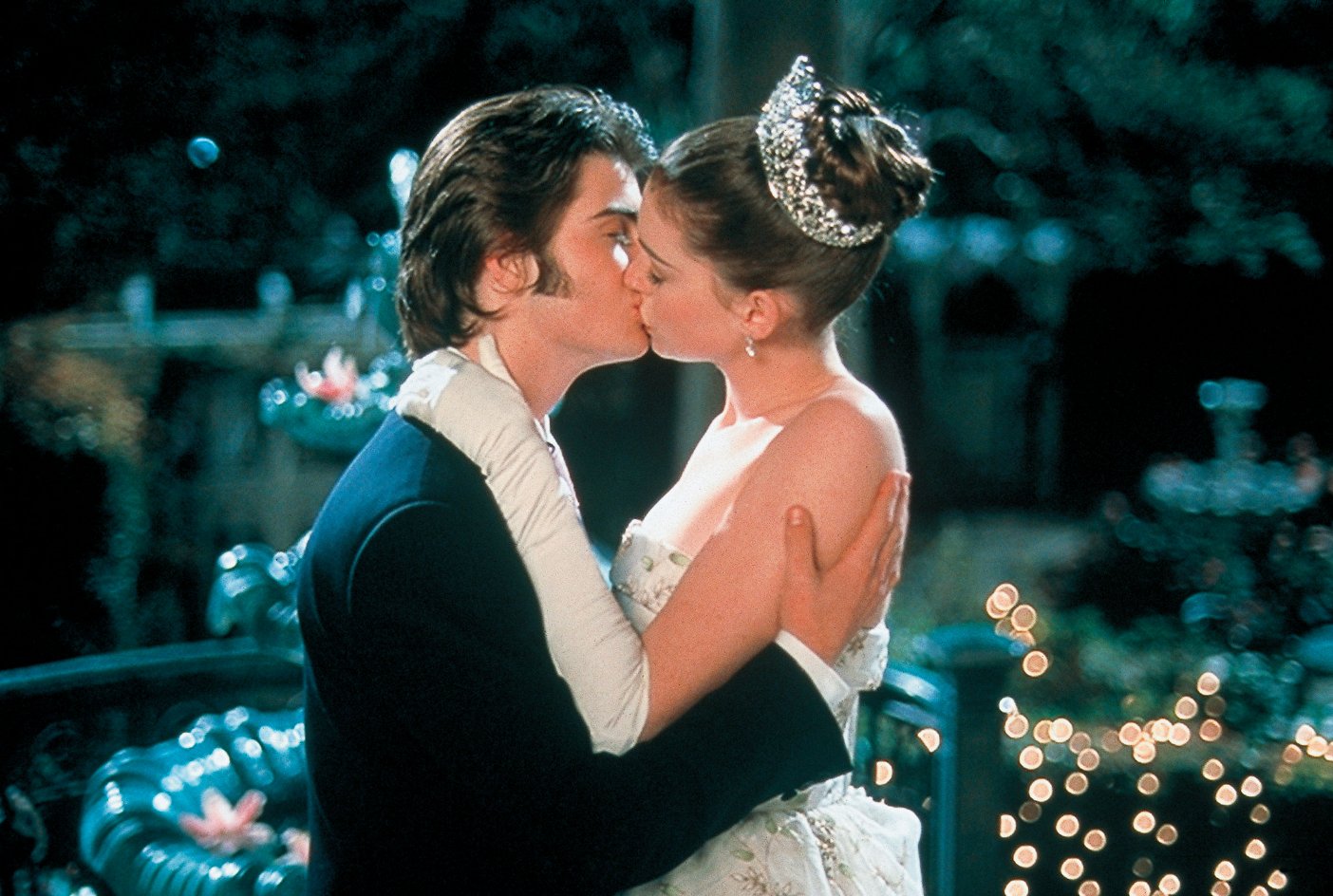}\hfill
    \includegraphics[height=3.2cm]{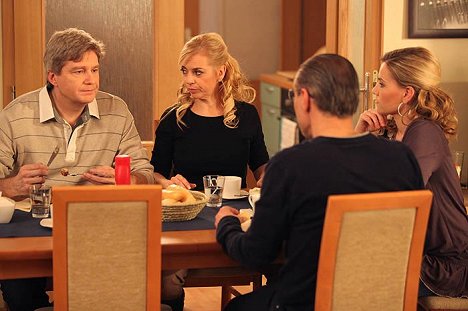}\hfill
    \includegraphics[height=3.2cm]{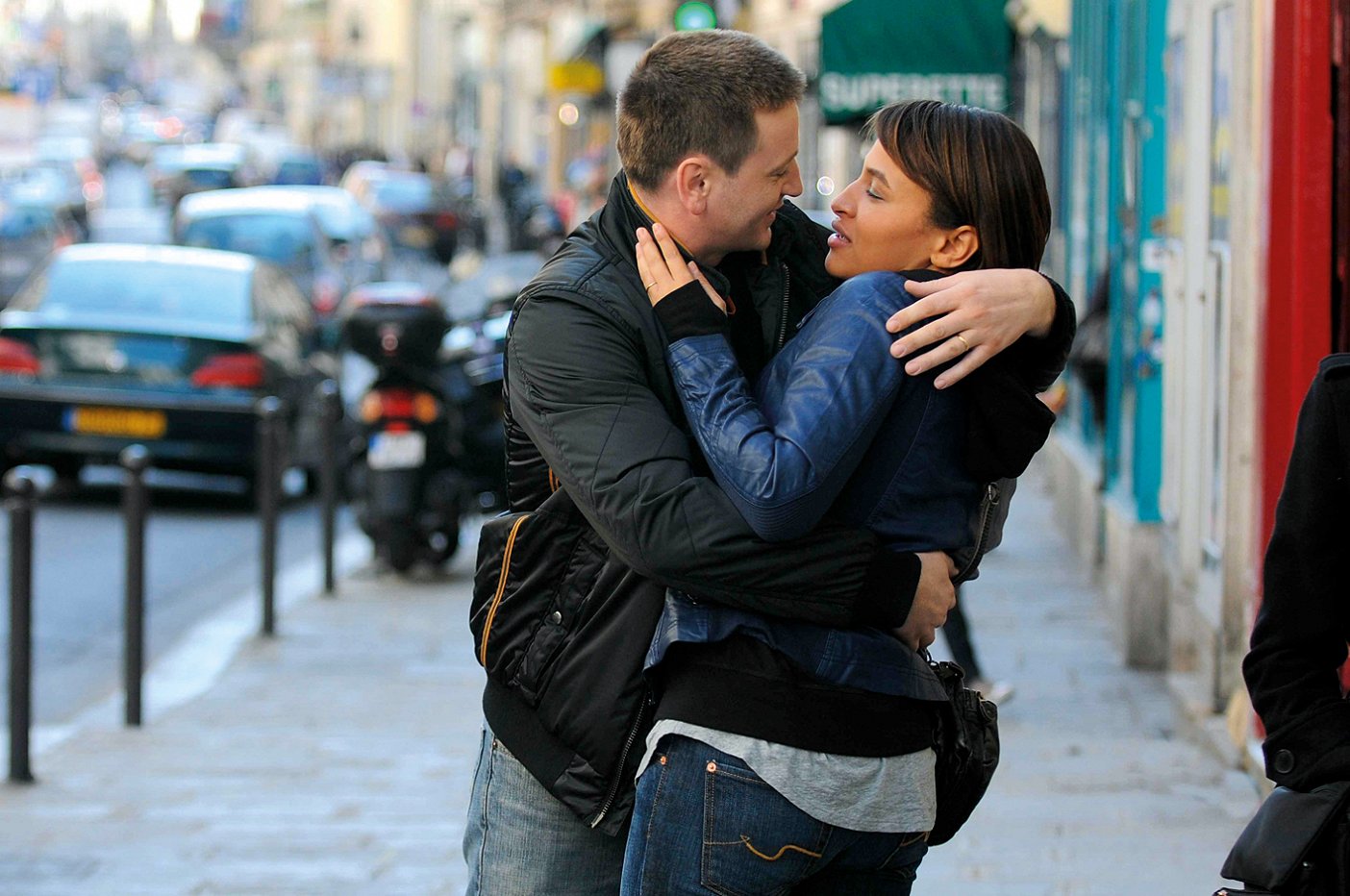}\hfill
    \includegraphics[height=3.2cm]{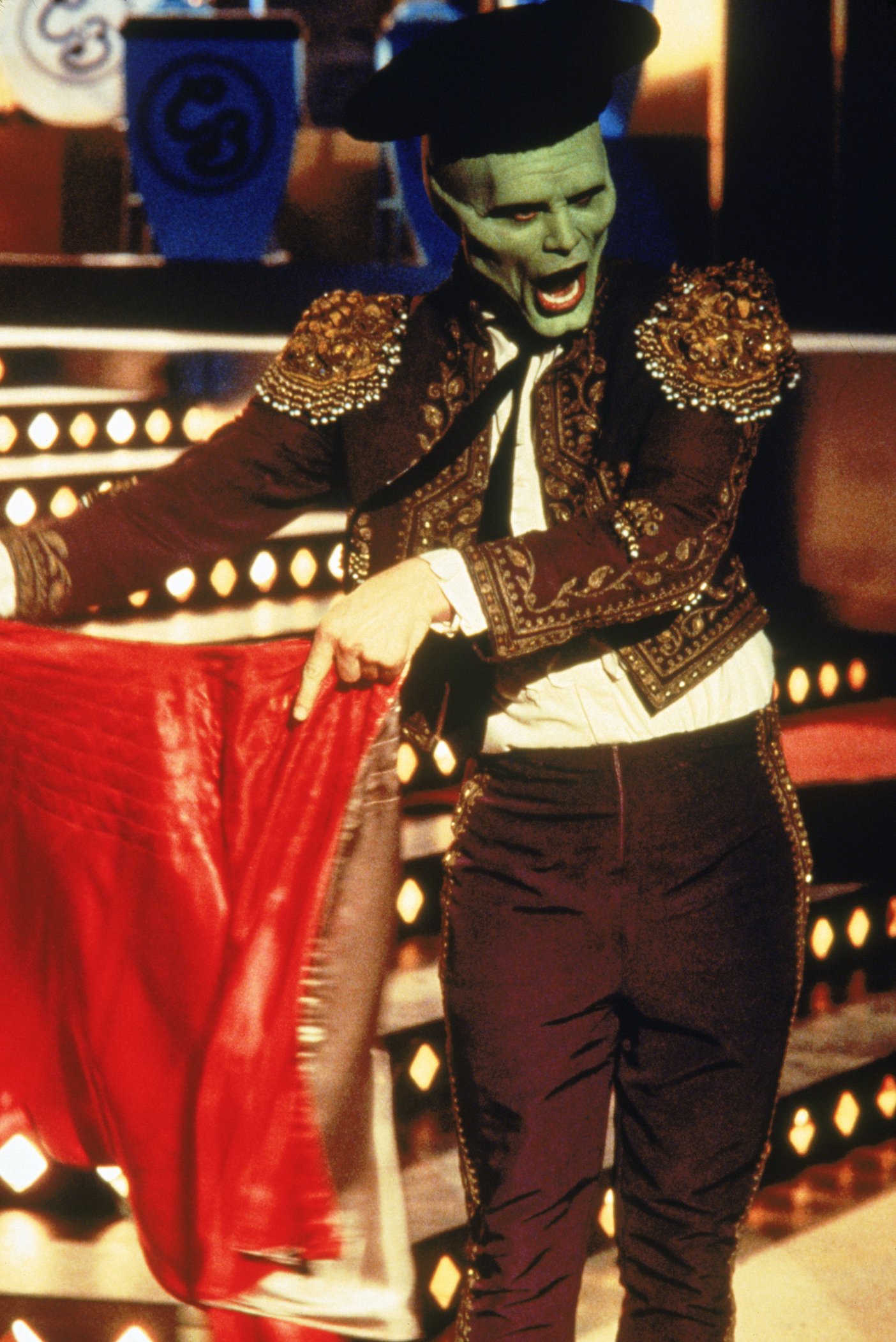}
    \caption{\textbf{Examples of Images With 0 Matches (Known Faces).}}
    \label{fig:images_with_no_matches}
\end{figure*}

It is crucial, however, to interpret the \textit{Scene} baseline performance as an optimistic, best-case scenario, as it was trained and evaluated on in-distribution data from CSFD-1.6M. The impact of this in-domain advantage is starkly evident when evaluating performance on images containing faces for which our pipeline found no matching identities. For images with at least one known identity, the \textit{Scene} baseline achieves a Mean Absolute Error (MAE) of 3.25 years. This error increases sharply to 5.35 years for images containing only unmatched faces (examples in \Cref{fig:images_with_no_matches}).

\begin{figure}
  \centering
  \includegraphics[width=\linewidth]{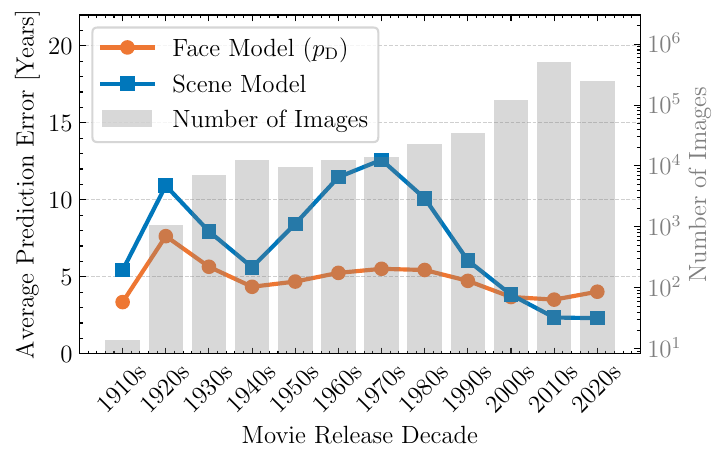}
    \caption{\textbf{Error by year analysis.} MAE $\downarrow$ of the {\textit{Face} ($p_{\text{D}}$)} and \textit{Scene} methods by the image capture year.}
    \label{fig:average_error_by_year}
\end{figure}

We hypothesize that the \textit{Scene} model learns not only from general temporal cues in the full image (e.g., fashion, color processing) but also implicitly learns to recognize frequently occurring actors and associate their specific appearance with a given time period. This dual reliance on the training distribution manifests in its failure modes. Performance degrades significantly when the model is confronted with faces that belong to less-known actors or extras. Moreover, as illustrated in \Cref{fig:average_error_by_year}, the model exhibits a strong temporal bias: its prediction error is lowest for years that are most frequent in the training set and highest for minority years. The ability of our proposed face-based model—which leverages external training data—to match and often outperform this powerful in-domain baseline is a testament to the potential of our approach.


\begin{figure}
  \centering
  \includegraphics[width=\linewidth]{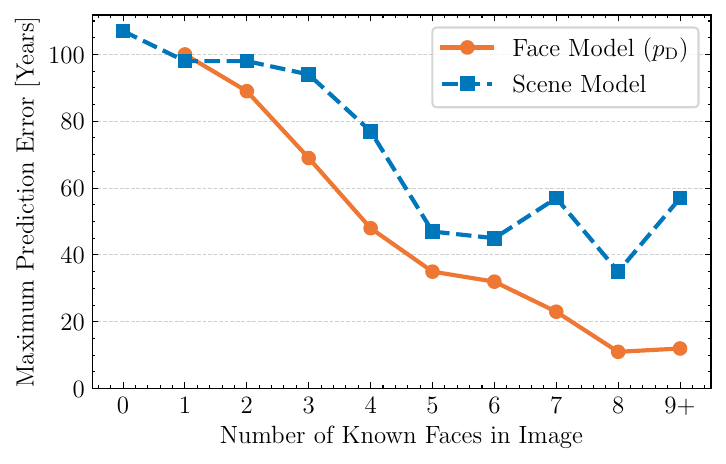}
    \caption{\textbf{Worst case error analysis.} Worst MAE $\downarrow$ of the {\textit{Face} ($p_{\text{D}}$)} and \textit{Scene} methods by the number of faces in the image.}
    \label{fig:worst_case_error_by_number_of_faces}
\end{figure}

\section*{CSFD-1.6M Quality vs. Cleaned IMDB-WIKI}
To distinguish the contribution of data quality from that of data quantity, we conducted a size-controlled experiment. We trained a ViT-B/16 age estimation model on a random subset of our dataset, referred to as CSFD-0.3M ($\approx$300,000 faces). To ensure the comparison was against the strongest possible baseline, we trained identical ViT-B/16 models on multiple cleaned versions of the IMDB-WIKI dataset, including the original \cite{Rothe-IJCV-2018}, IMDB-CLEAN \cite{9733166}, and the top-performing EM-CNN version \cite{Franc-IVC-2018}. We selected the EM-CNN version ($\approx$310,000 faces) for the definitive comparison, as it is both the strongest baseline and the most comparable in size to our subset.

The results are presented in \Cref{tab:age_estimation_results_new}. The model trained on CSFD-0.3M consistently and significantly outperforms this best-performing IMDB-WIKI-trained model across all evaluation benchmarks. Since the training set sizes are directly comparable, this performance gap provides strong empirical evidence that CSFD-1.6M contains a substantially higher-quality training signal for age estimation, validating its contribution beyond mere scale.

\begin{table}
\centering
\scriptsize
\begin{tabular}{@{}l cccc@{}}
\toprule
& \multicolumn{4}{c}{\textbf{Pre-training Dataset}} \\
\cmidrule(l){2-5} 
\textbf{Benchmark} & \textbf{ImageNet} & \textbf{IMDB-WIKI} & \textbf{CSFD-0.3M} & \textbf{CSFD-1.6M}
\\
\midrule
AgeDB      & 7.05 $\pm$ 0.29 & 6.34 $\pm$ 0.25 & 5.46 $\pm$ 0.24 & \textbf{5.25 $\pm$ 0.21} \\
AFAD       & 3.19 $\pm$ 0.04 & 3.10 $\pm$ 0.03 & 3.08 $\pm$ 0.01 &
\textbf{3.04 $\pm$ 0.03} \\
MORPH      & 2.98 $\pm$ 0.05 & 2.88 $\pm$ 0.07 & 2.83 $\pm$ 0.07 & \textbf{2.76 $\pm$ 0.05} \\
UTKFace    & 4.84 $\pm$ 0.08 & 4.64 $\pm$ 0.06 & 4.23 $\pm$ 0.02 & \textbf{4.08 $\pm$ 0.03} \\
CLAP2016   & 5.87 & 4.89 & 3.53 & \textbf{3.52} \\
\bottomrule
\end{tabular}
\caption{\textbf{Age estimation.} MAE $\downarrow$ ($\pm$ std) after pre-training a ResNet-101 model on different datasets. The CSFD-0.3M is size-matched to the cleaned version of IMDB-WIKI, EM-CNN \cite{Franc-IVC-2018}.}
\label{tab:age_estimation_results_new}
\end{table}

\begin{figure*}[!hb]
  \centering
    \begin{subfigure}{0.3\linewidth}
    \includegraphics[width=\linewidth]{figures/yearly_distribution.pdf}
    \caption{Temporal distribution of the data}
    \label{fig:short-c}
  \end{subfigure}
  \hfill
    \begin{subfigure}{0.3\linewidth}
    \includegraphics[width=\linewidth]{rebuttal_figures/dist_faces_per_image_with_zeros.pdf}
    \caption{Distribution of known faces per image}
    \label{fig:short-b}
  \end{subfigure}
  \hfill
  \begin{subfigure}{0.3\linewidth}
    \includegraphics[width=\linewidth]{figures/age_distribution.pdf}
    \caption{Distribution of face ages}
    \label{fig:short-a}
  \end{subfigure}\\
  \centering
    \begin{subfigure}{0.3\linewidth}
    \includegraphics[width=\linewidth]{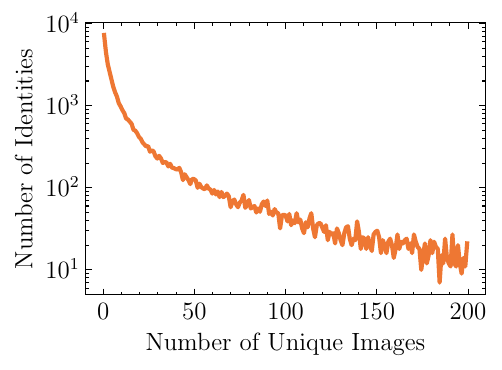}
    \caption{Distribution of unique images per identity}
    \label{fig:short-c}
  \end{subfigure}
  \hfill
    \begin{subfigure}{0.3\linewidth}
    \includegraphics[width=\linewidth]{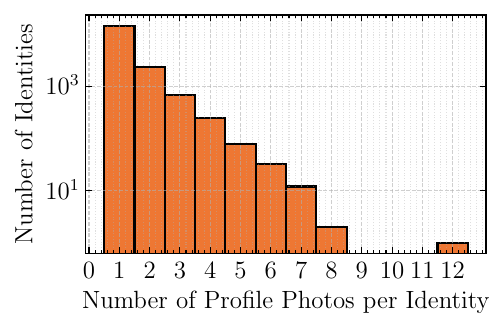}
    \caption{Distribution of profile images per identity}
    \label{fig:short-b}
  \end{subfigure}
  \hfill
  \begin{subfigure}{0.3\linewidth}
    \includegraphics[width=\linewidth]{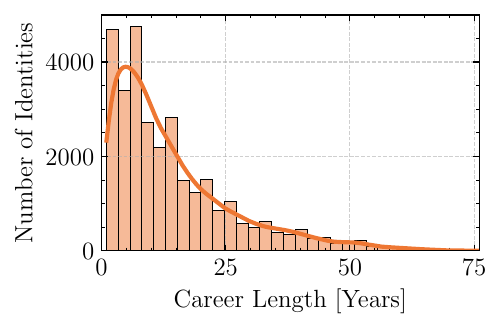}
    \caption{Distribution of active career lengths}
    \label{fig:short-a}
  \end{subfigure}  
  \caption{\textbf{Overview of the CSFD-1.6M dataset statistics.}}
  \label{fig:dataset_statistics_supplementary}
\end{figure*}

\section*{Scalability and Computational Cost}
The proposed method is scalable. The initial \textit{feature extraction}, involving the computation of face embeddings and age posteriors, represents a one-time cost and is highly parallelizable on GPU hardware. The dating logic, which includes the similarity search and marginalization over assignments, is performed on a CPU. In our experiments, the dating process on a single CPU core achieved an average inference speed of 0.5~seconds per image, a figure that includes the computation for both the \textit{Full} and \textit{Top-1} models. Since the dating of each image is an independent task, the overall throughput can be significantly increased for batch processing via standard multiprocessing. The primary computational bottleneck is the similarity search against the $\approx$46,000 identities in our database. For intended applications such as dating genealogical photographs, the set of known identities would be substantially smaller, leading to a considerable reduction in inference time.

\section*{Qualitative Analysis of Failure Cases}
\begin{figure*}
  \centering

  \begin{subfigure}{\linewidth}
    \centering
    \includegraphics[height=2.5cm]{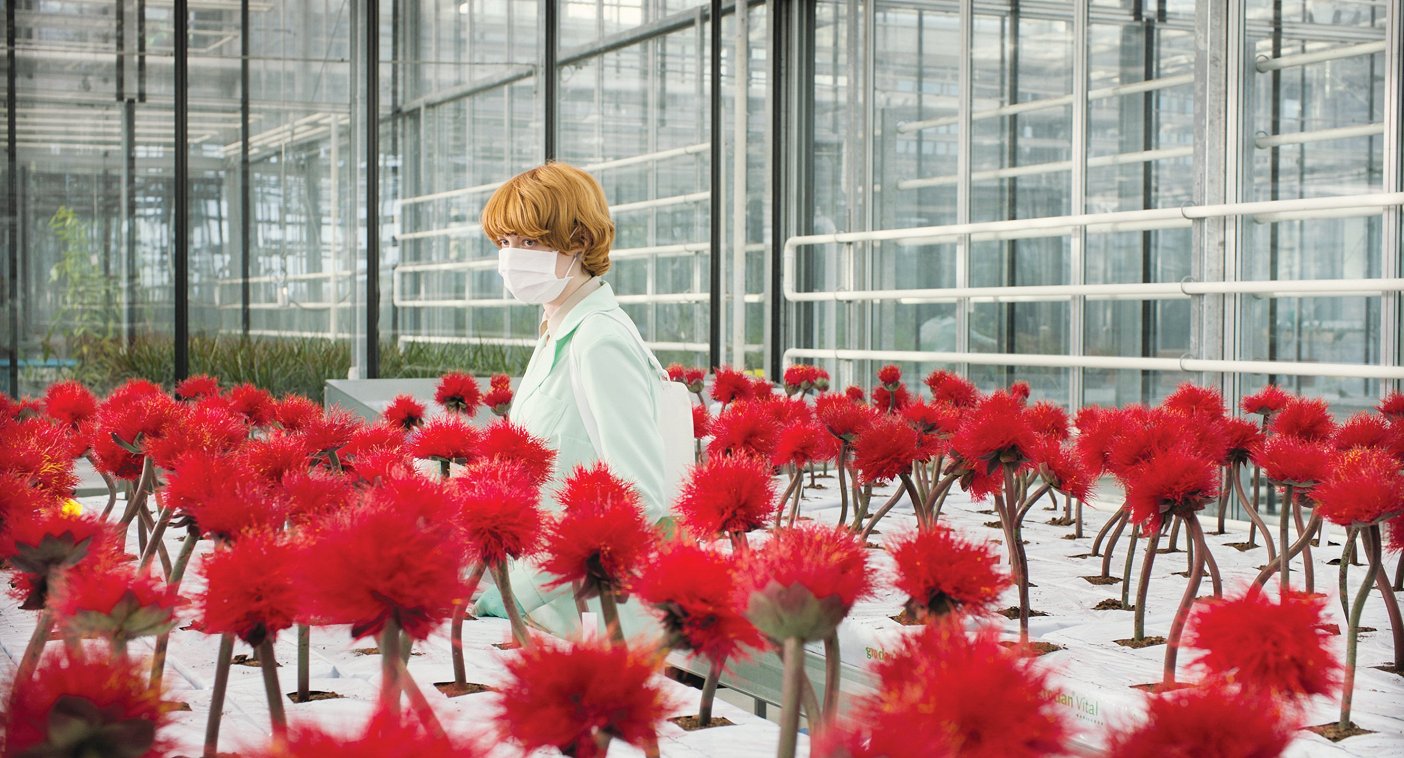}\hfill
    \includegraphics[height=2.5cm]{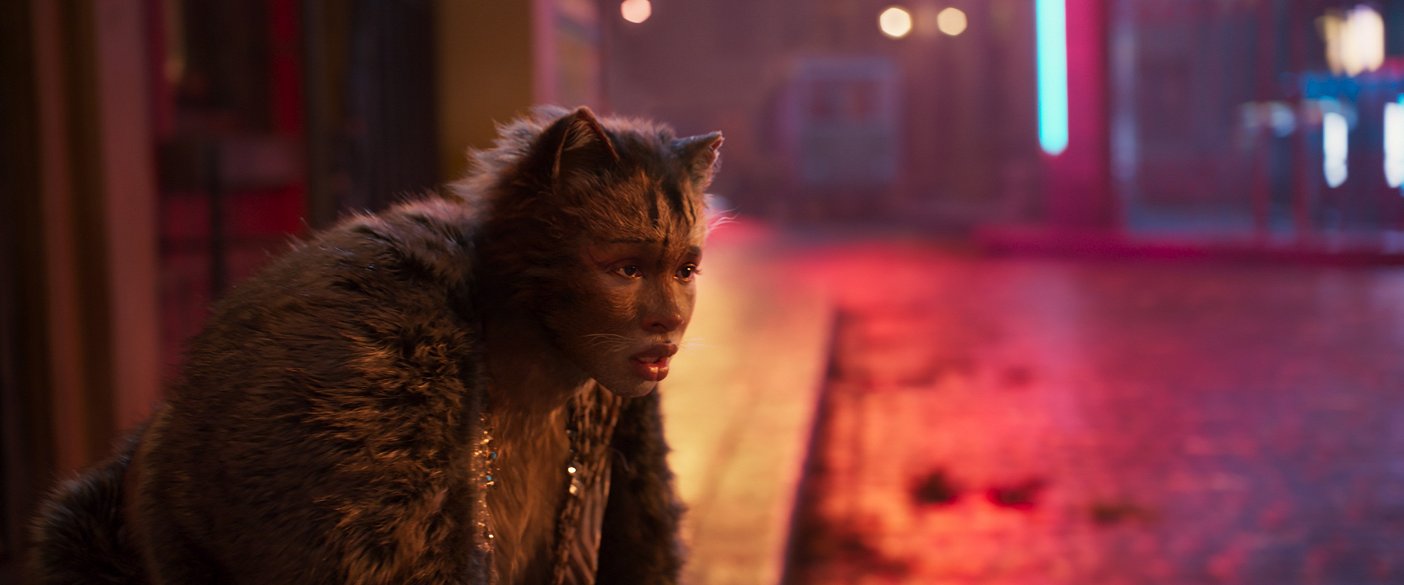}\hfill
    \includegraphics[height=2.5cm]{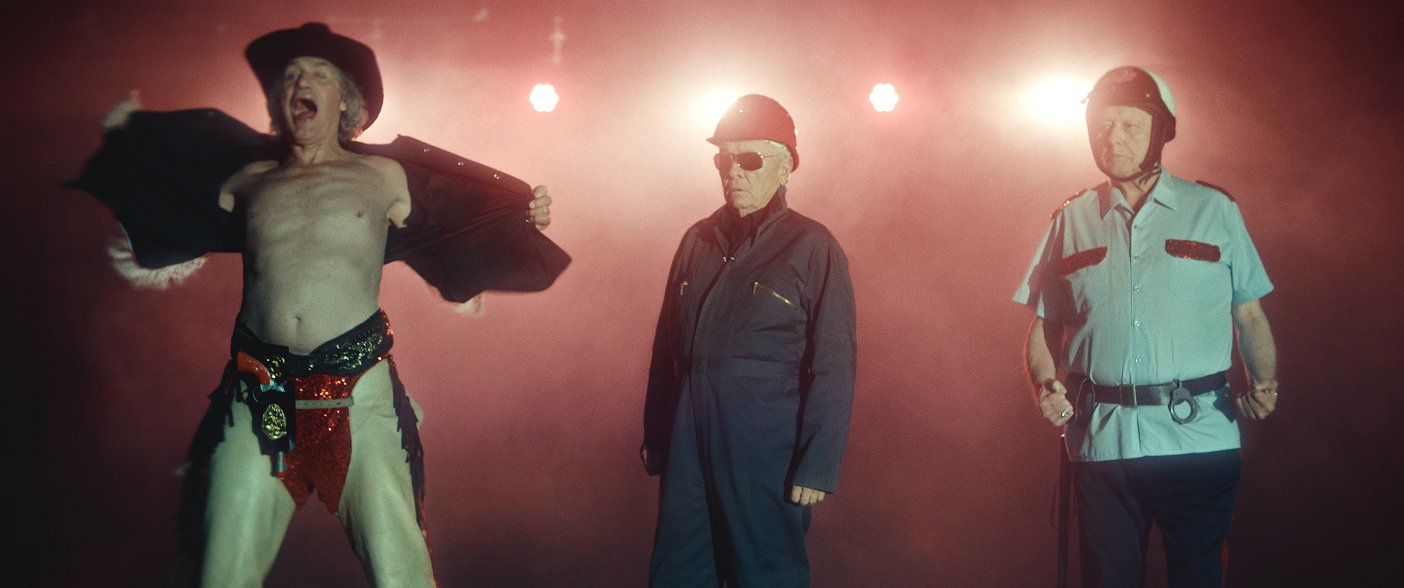}
    \caption{Failure cases caused by incorrect identification due to masks, heavy makeup or extreme expressions.}
    \label{fig:no_matches}
  \end{subfigure}

  \vspace{0.6cm}

  \begin{subfigure}{\linewidth}
    \centering
    \includegraphics[height=2.9cm]{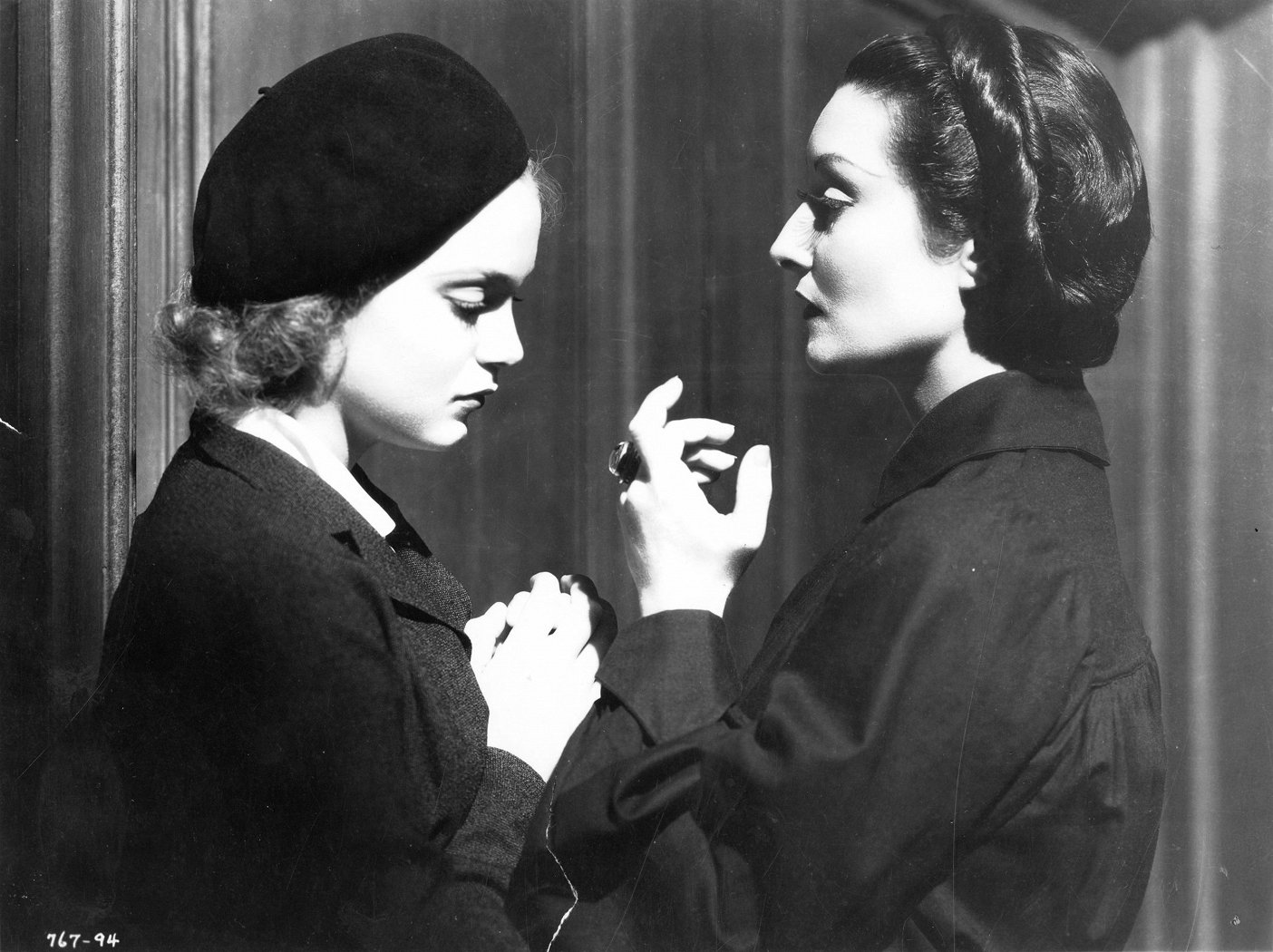}\hfill
    \includegraphics[height=2.9cm]{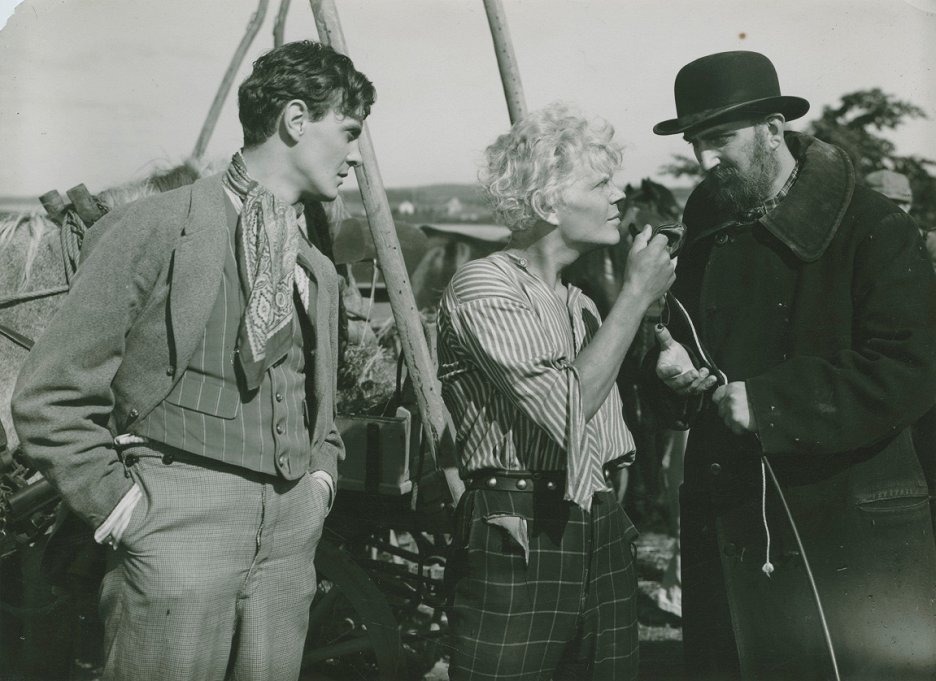}\hfill
    \includegraphics[height=2.9cm]{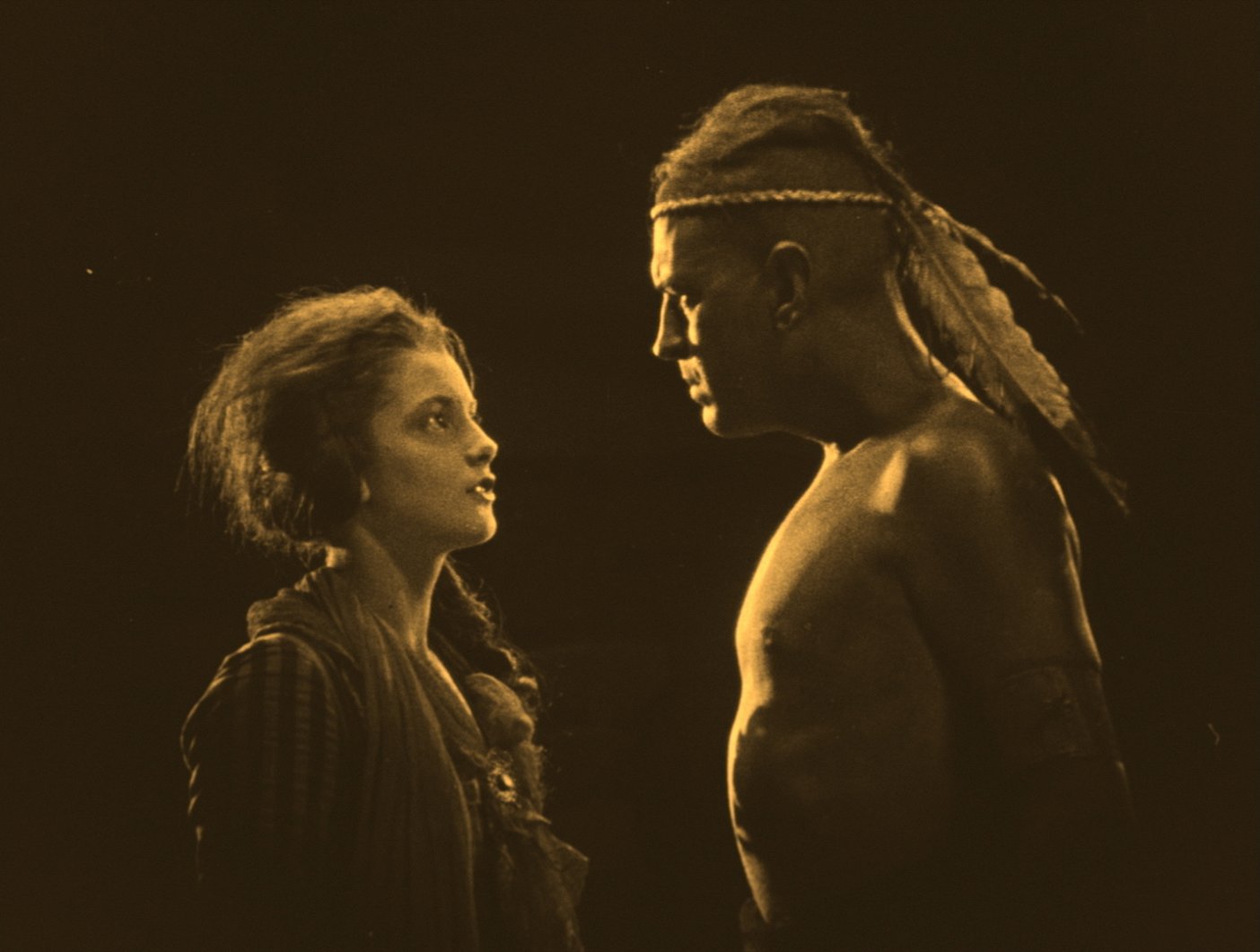}\hfill
    \includegraphics[height=2.9cm]{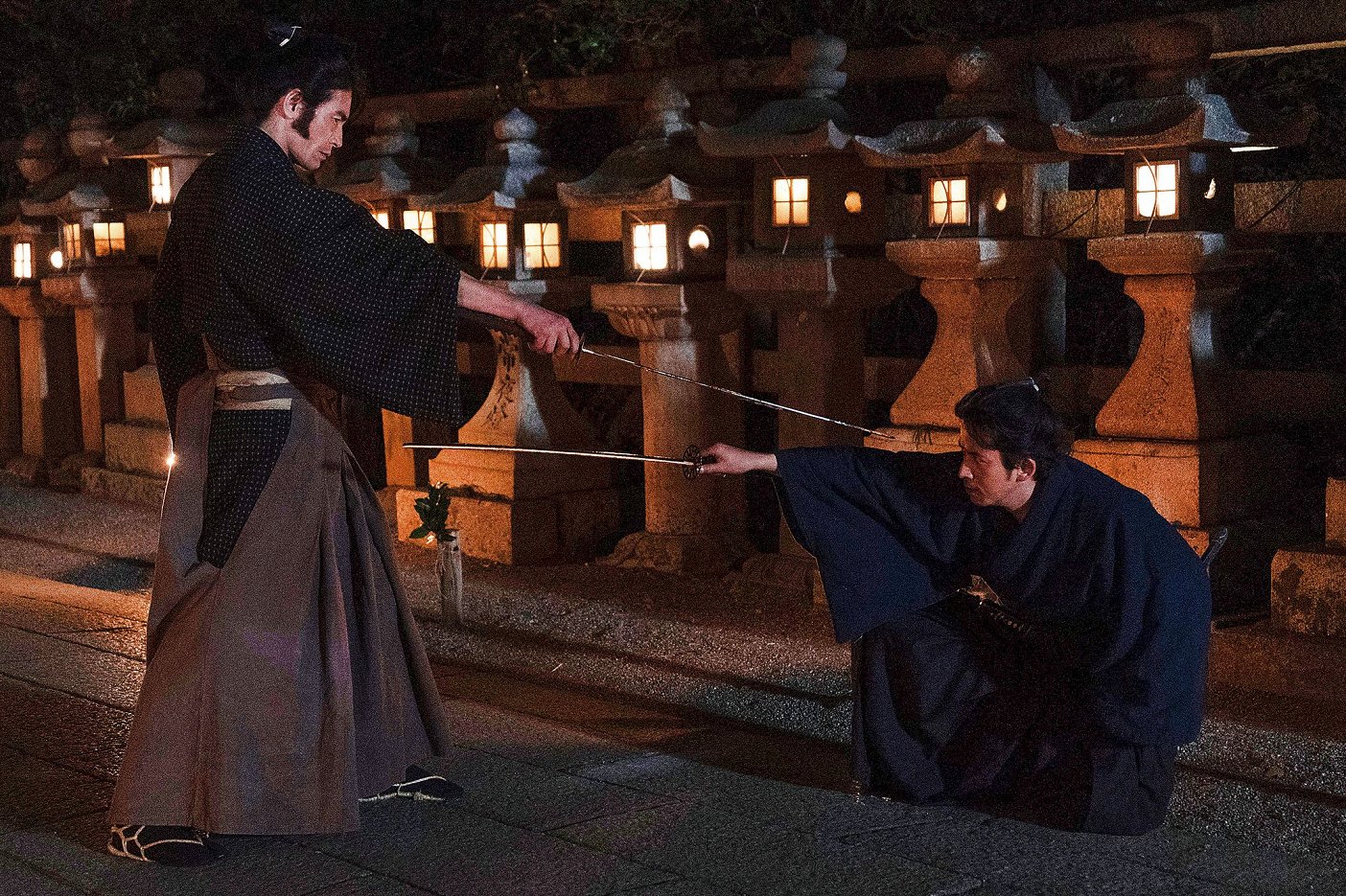}
    \caption{Failure cases caused by incorrect identification due to actors being viewed from the side.}
    \label{fig:with_matches}
  \end{subfigure}

  \caption{\textbf{Examples of Dating Failure Cases.}}
  \label{fig:face_dating_failures}
\end{figure*}

A key advantage of the proposed facial-age-based model is the interpretability of its failure modes. It is possible to differentiate whether a prediction error originates from an incorrect identity assignment or from an inaccurate age estimation. An analysis of the highest MAE failure cases, presented in \Cref{fig:face_dating_failures}, reveals that errors predominantly stem from the face recognition component. Recognition failures are typically caused by challenging conditions such as masks, glasses, heavy makeup, extreme facial expressions, or difficult poses including profiles and long-distance shots.

In contrast, the failure cases of the \textit{Scene} model are not readily interpretable due to its black-box nature. However the largest prediction errors correlate strongly with the temporal distribution of the training data. As visualized in \Cref{fig:average_error_by_year}, the model performs best on years that are most represented in the dataset. Further visualizations of the error characteristics for both models, stratified by the number of known faces, are provided in \Cref{fig:worst_case_error_by_number_of_faces}.

\section*{Ethical Considerations} 
We release our contributions to foster research in temporal media analysis and encourage their responsible application, however, we recognize the associated ethical duties. The use of public data scraped without explicit consent, the demographic bias from the Czecho-Slovak Movie Database, and the potential for dual-use are significant concerns. To mitigate these while supporting reproducible research, we don't distribute the original images, but only the annotations, pre-computed features and URLs.

\begin{figure*}
    \centering
    \includegraphics[width=\linewidth]{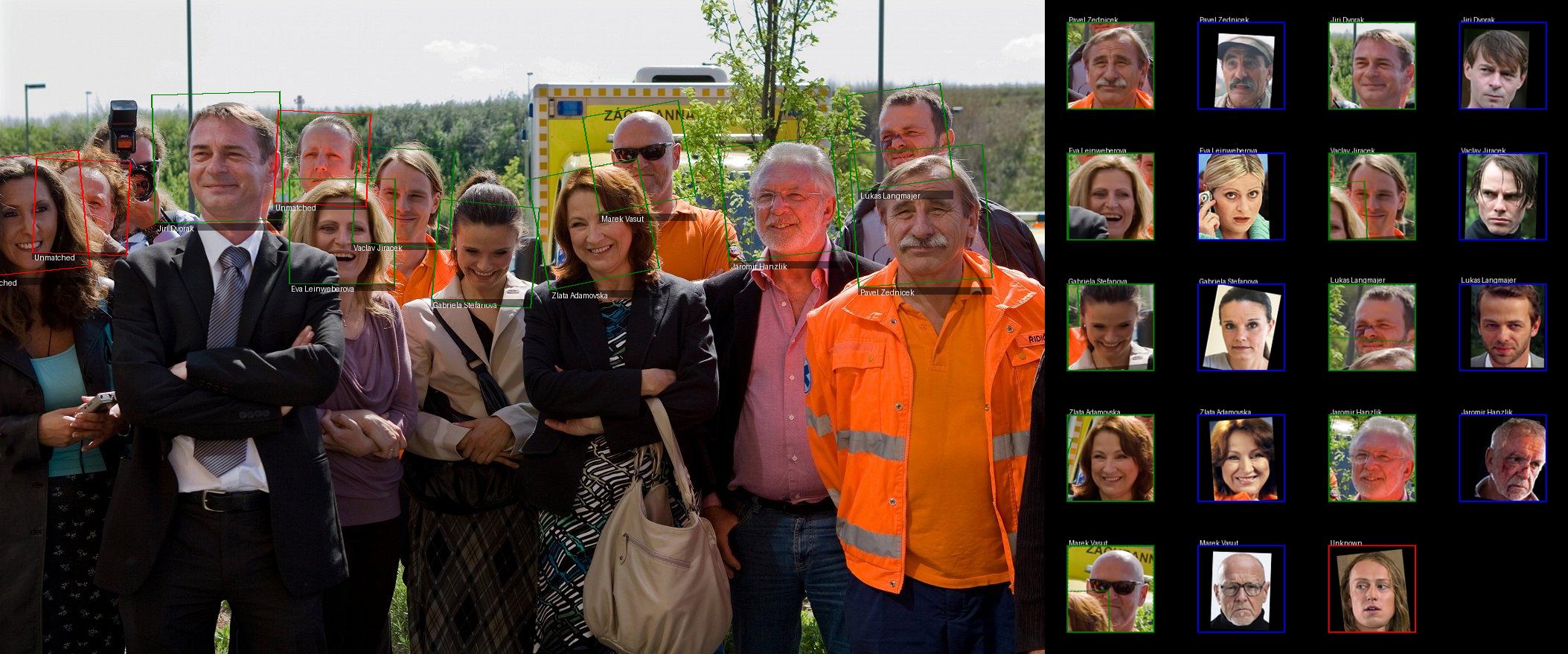}\\
    \includegraphics[width=\linewidth]{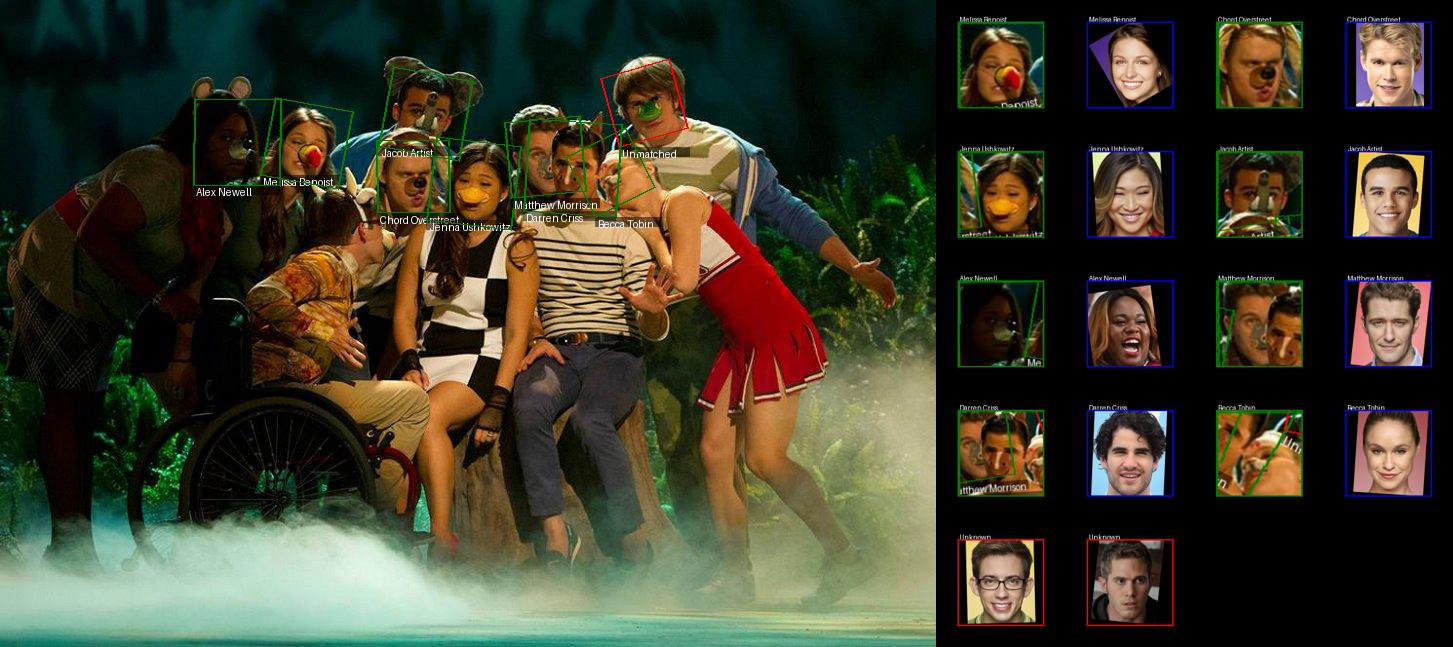}
    \caption{\textbf{Examples of Face Matching Results.}}
    \label{fig:image_matches_examples}
\end{figure*}
\begin{figure*}
    \centering
    \includegraphics[width=\linewidth]{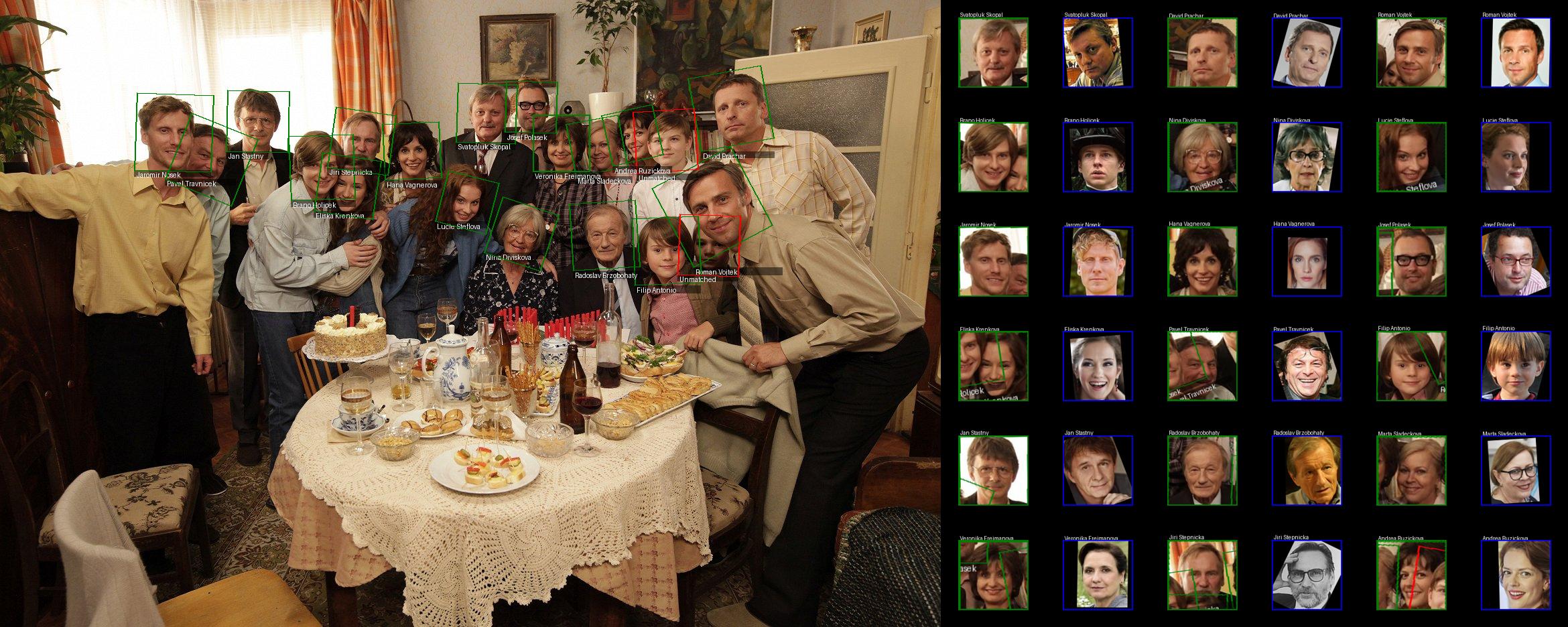}\\
    \includegraphics[width=\linewidth]{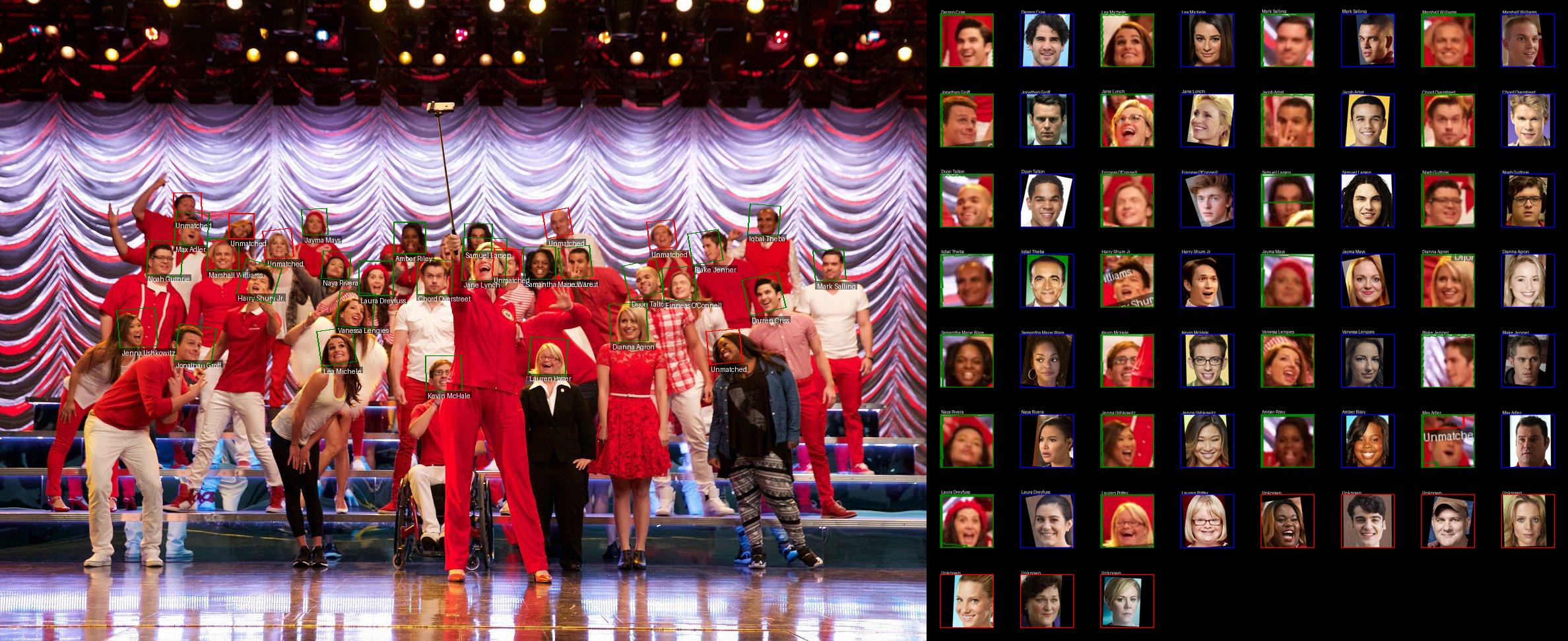}\\
    \includegraphics[width=\linewidth]{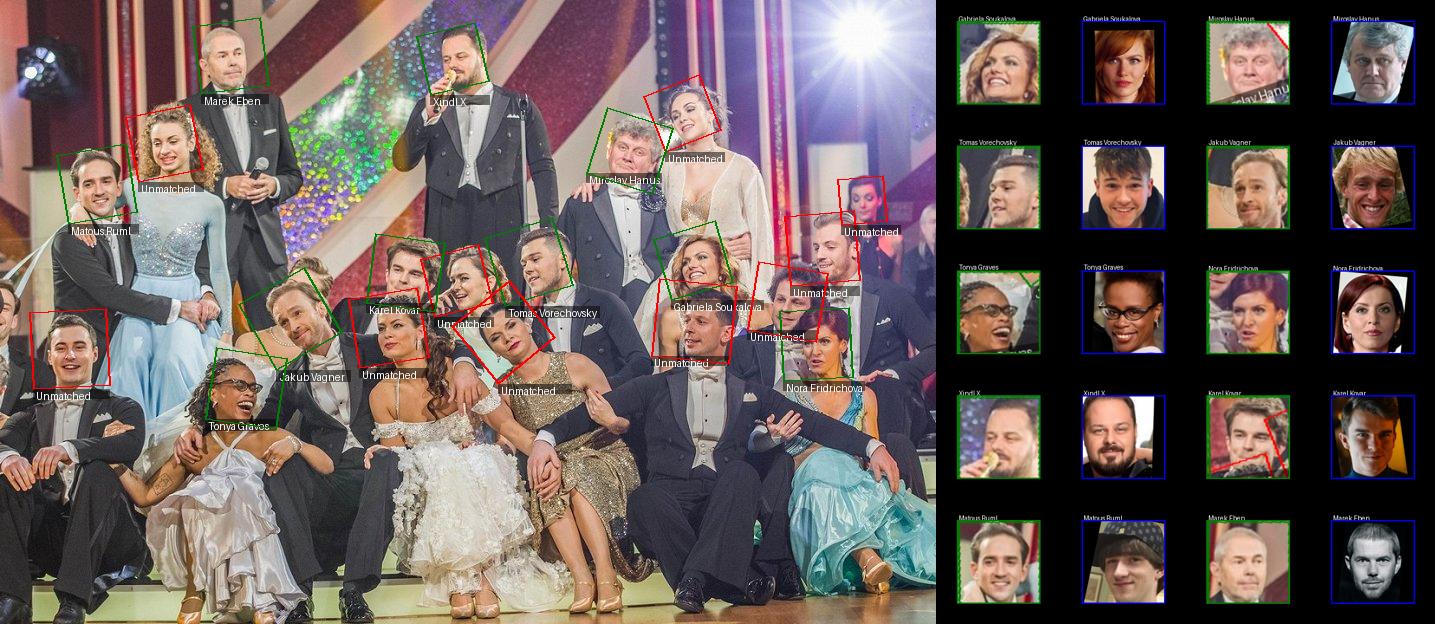}
    \caption{\textbf{Examples of Face Matching Results.}}
    \label{fig:image_matches_examples}
\end{figure*}
\begin{figure*}
    \centering
    \includegraphics[width=\linewidth]{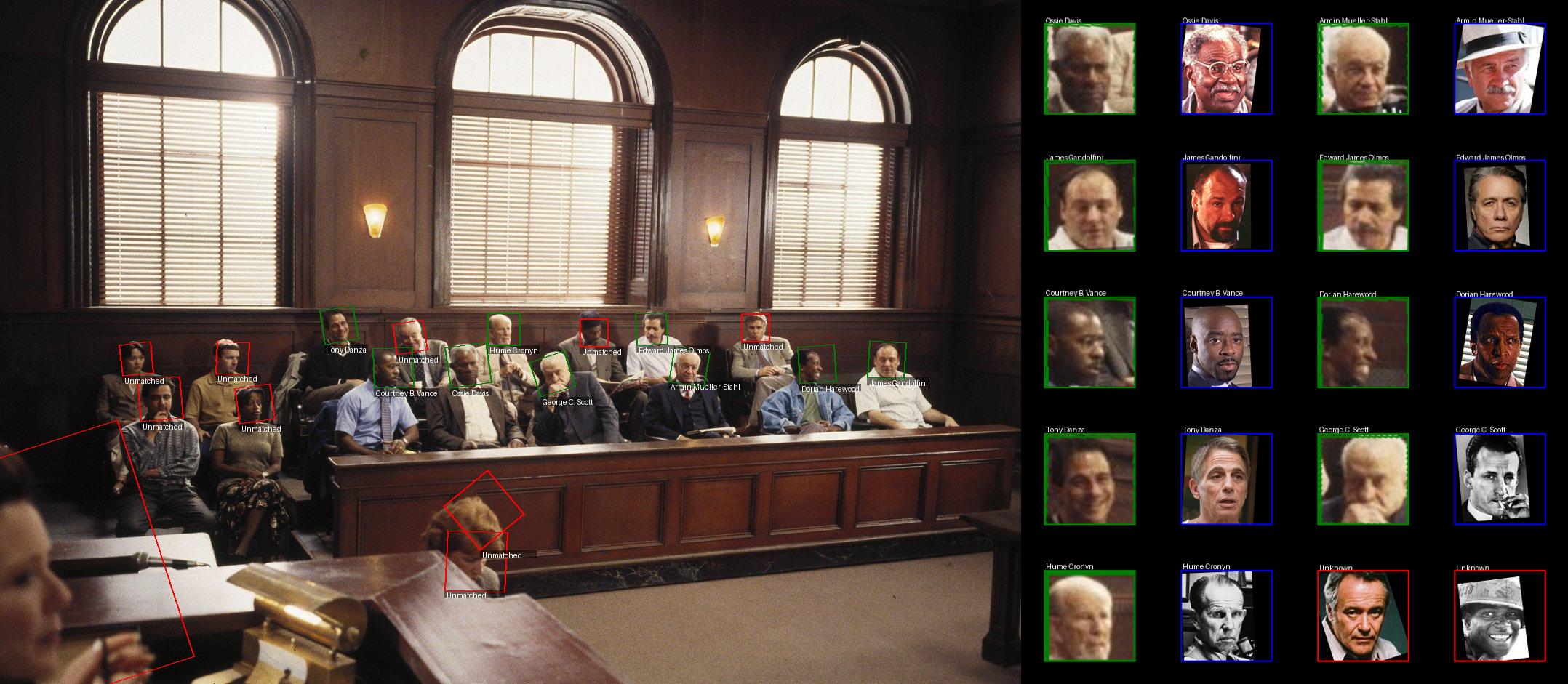}\\
    \includegraphics[width=\linewidth]{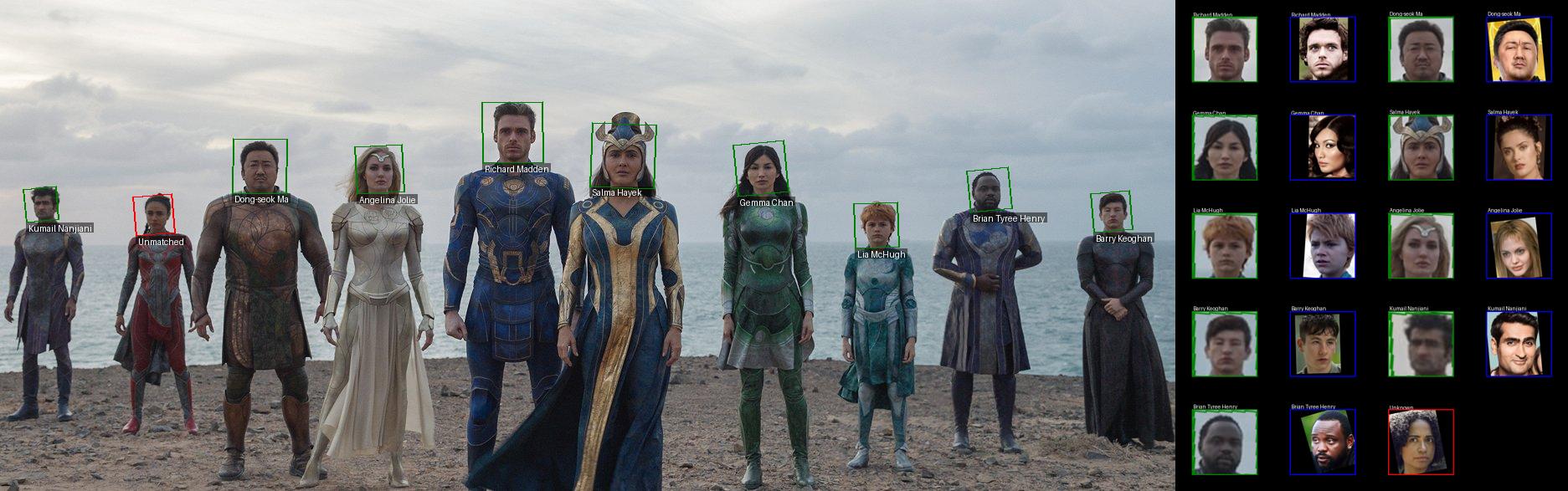}
    \caption{\textbf{Examples of Face Matching Results.}}
    \label{fig:image_matches_examples}
\end{figure*}

\end{document}